# Adopting a human developmental visual diet yields robust and shape-based AI vision

Zejin Lu*†[1,2], Sushrut Thorat*[1], Radoslaw M Cichy[2], & Tim C Kietzmann†[1]

[1] Machine Learning Group, Institute for Cognitive Science, Osnabrück University, Osnabrück, Germany.
[2] Neural Dynamics of Visual Cognition Group, Department of Education and Psychology, Freie Universität Berlin, Berlin, Germany.
* Shared first authorship.
† Corresponding author

## Abstract

Despite years of research and the dramatic scaling of artificial intelligence (AI) systems, a striking misalignment between artificial and human vision persists. Contrary to humans, AI relies heavily on texture-features rather than shape information, lacks robustness to image distortions, remains highly vulnerable to adversarial attacks, and struggles to recognise simple abstract shapes within complex backgrounds. To close this gap, here we take inspiration from how human vision develops from early infancy into adulthood. We quantified visual maturation by synthesising decades of research into a novel developmental visual diet (DVD) for AI vision. Guiding AI systems through this human-inspired curriculum, which considers the development of visual acuity, contrast sensitivity, and colour, produces models that better align with human behaviour on every hallmark of robust vision tested, yielding the strongest reported reliance on shape information to date, abstract shape recognition beyond the state of the art, and higher resilience to image corruptions and adversarial attacks. Our results thus demonstrate that robust AI vision can be achieved by guiding how a model learns, not merely how much it learns, offering a resource-efficient route toward safer and more human-like artificial visual systems.

## Main

Despite the unprecedented scale and capabilities of contemporary AI vision systems, they remain fundamentally misaligned with human vision[2,8]. One striking misalignment is that, unlike humans, AI vision systems continue to rely on texture-based rather than shape-based features[1–3]. Relatedly, AI vision exhibits poor generalisation to image degradations[4], is often unable to recognise abstract shapes[8], and suffers from adversarial attacks[5,6,9]. These shortcomings constitute an important roadblock in the development of reliable, human-aligned AI.

In aiming to solve these issues, we here subscribe to the hypothesis that the key to human-aligned AI vision lies in overcoming the fundamental differences in visual upbringing between humans and AI systems[10–12]. While AI systems are trained with high-fidelity input throughout their training trajectory, human visual development is, initially, severely limited in aspects such as visual acuity. Evidence for the functional importance of a gradual visual development can be found in both neuroscience and AI-based modelling. For example, children whose congenital cataracts are removed can nevertheless exhibit lasting configuration recognition deficits[13], likely due to missing parts of the slow developmental trajectory of healthy controls. Accordingly, AI systems that involve training with fixed blurring stages were shown to be more robust[14–19], albeit remaining far from human-level shape bias and robustness to image degradations. Notably, these modelling studies focused on visual acuity development alone, abstractly modelled as one, or up to a few discrete bins i.e. not reflecting the continuous developmental trajectory of human vision. Building on these initial experiments, we here aim to test whether AI vision can benefit further from bringing its visual diet into closer agreement with the visual developmental trajectory of humans, while considering developmental factors beyond acuity.

To model the continuous development of human vision, we reviewed the existing psychophysical evidence on how human vision develops from newborn to 25 years of age and synthesised the developmental trajectories of three core dimensions of visual maturation (visual acuity[20–34], chromatic sensitivity[35–38], and contrast sensitivity[39–45]) into a preprocessing pipeline for AI vision. Across a series of experiments, we demonstrate that guiding AI through this human developmental visual diet (DVD), instead of the gold standard of high-resolution AI vision training, is an effective means towards approaching human-like robustness in visual inference. Deep neural networks (DNNs) trained with DVD exhibit shape bias beyond the state of the art, greater reliance on spatial integration of visual features, increased capability of abstract shape recognition, as well as heightened resilience to image degradations and adversarial attacks. Importantly, the translation of human psychophysical data into an efficient image preprocessing pipeline for AI necessarily involves selecting a level of abstraction that balances replication detail with algorithmic considerations that will enable easy adoption in the field. Due to the many remaining differences with biological vision, DVD is thus not to be considered a veridical mirror of the perceptual world of infants.

## Results

In our experiments, we contrasted DNNs trained on high-res images (gold standard) with networks trained with DVD, a developmental trajectory set up so that the full model training corresponds to 25 years of human visual development. Various model architectures were trained with an object categorization objective across multiple image datasets, including mini-ecoset[46], ecoset[47] or ImageNet-1K[48]. Subsequent to training, the models were evaluated across a test battery for inferential robustness (Fig. 1a), comparing models to human behavioural data where available.

The mapping from chronological months to training epochs was handled by the hyperparameter $\alpha$. For the development of visual acuity, we mapped the Snellen-equivalent terms onto a Gaussian blur parameter ($\sigma$, Fig. 1b; see Methods). Contrast sensitivity trajectories were converted into a frequency-domain threshold modulation (Fig. 1c; see Methods), controlled by hyperparameters $\beta$ (initial threshold) and $\lambda$ (sensitivity mapping). Chromatic sensitivity was implemented via adjustments in colour fidelity (Fig. 1d; see Methods). By exploring the three hyperparameters of DVD ($\alpha$, $\beta$, $\lambda$), we modulated the extent and intensity of early visual experiences the model receives during training (see Fig. 1e for example trajectories). Irrespective of model training setup or training age, all testing was performed on the same high-res test images.

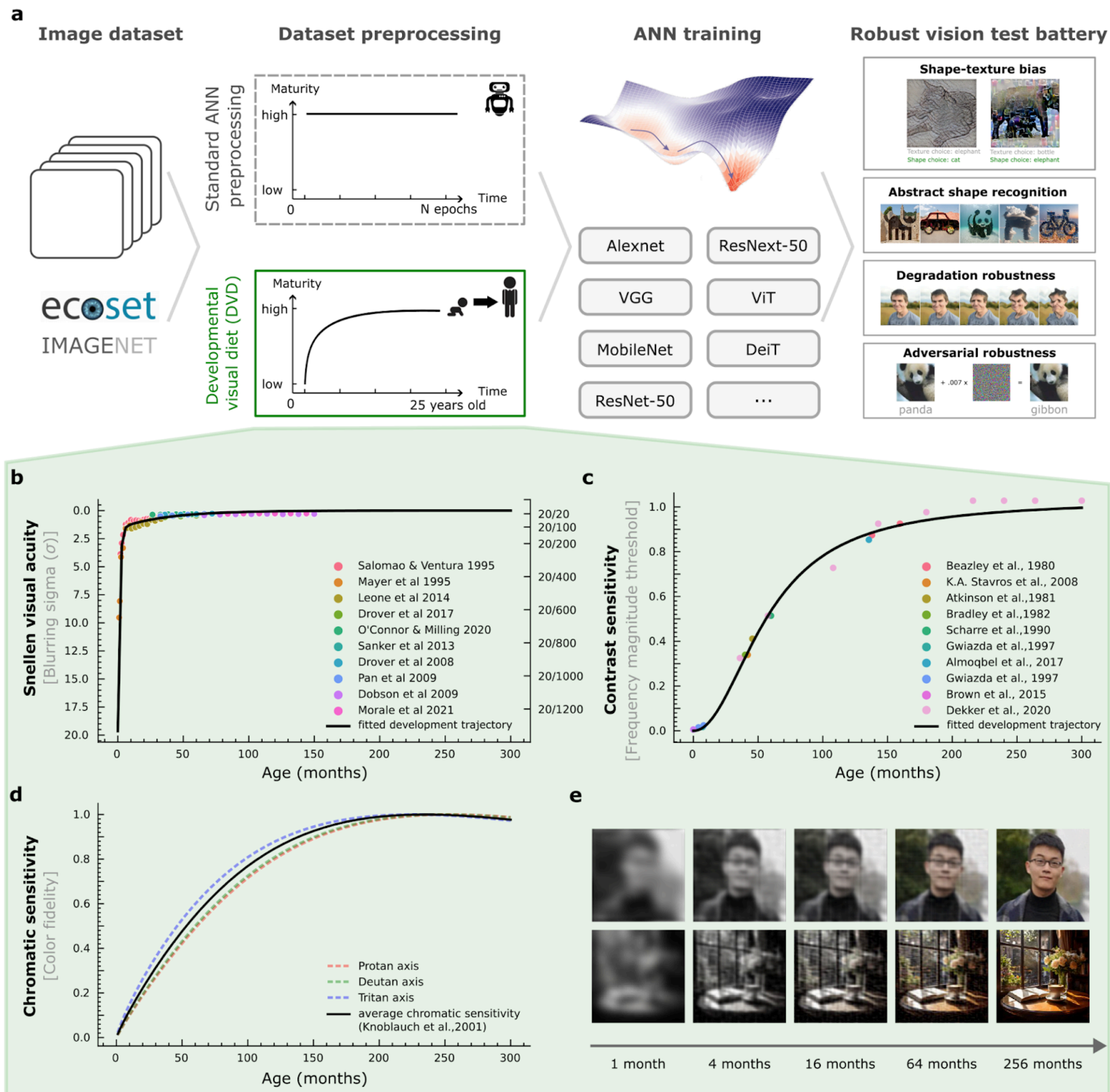

**Figure 1 | Following the developmental trajectory of visual acuity, chromatic sensitivity, and contrast sensitivity from newborn to 25-year-old, as an effective means of training robust AI vision systems.** ***a****. Schematic of our training and analysis pipeline for comprehensively evaluating Developmental Visual Diet (DVD) vs standard training regimes for artificial neural networks (ANNs).* ***b-d****. Developmental trajectories of the three aspects of vision modelled by DVD, as synthesised from a multitude of psychophysical experiments across age groups. In our main experiments, we represent contrast by the peak sensitivity across spatial frequencies at each age; but see Fig. S6 for results based on full frequency-dependent developmental profiles, and see Fig. S14 for results based on contrast sensitivity function (CSF) multiplication and frequency pooling (see Methods).* ***e****. DVD example images at selected developmental stages, illustrating the progression of visual experience from newborns to adults.*

### ***DVD training promotes near-human levels of shape bias***

A defining feature of human visual perception is that, when presented with object images for which the texture and shape information are in conflict (Fig. 2a), observers reliably report perceiving the object category in line with the shape information (their shape bias score is around 0.96 (96%) in adults[1] and 0.90 (90%) in 4-year-olds[49]). By contrast, DNNs, irrespective of network architecture, training dataset, dataset size, or training objective, exhibit a strong tendency towards texture-based decisions (their shape bias scores typically fall in the 0.2-0.4 range[1,50]). The texture-preference in DNNs therefore constitutes a signature difference to human vision.

We first explored the effect of DVD on the shape/texture bias in ANNs by performing a comprehensive hyperparameter sweep of DVD based on the ResNet-50 model architecture[51], trained on a smaller-scale image-dataset (mini-ecoset[46], 280k images, for training details see Methods, Fig. 2b). Observing a tradeoff between shape-selectivity and accuracy across hyperparameter settings, three representative model configurations were chosen for subsequent analyses: shape-bias favoured (DVD-S; 0.94 shape bias, aligning closely with the human range of 0.90-0.97), recognition performance favoured (DVD-P; increased shape bias from 0.32 (baseline) to 0.73 while getting similar overall accuracy to baseline model), and striking a balance between the two (DVD-B; 0.82 shape bias with a modest accuracy loss of approximately 4.9% compared to gold-standard training with high-res images). Crucially, when scaling these 3 variants to full ecoset[47] training (1.5 million images, 565 basic-level categories), the prior results were closely reproduced (Fig. 2b, right). While the control model achieved a shape bias of 0.34 at 62.99% accuracy, DVD-S delivered a shape bias of 0.90, i.e. within the human range; DVD-P exhibited a shape bias of 0.70, alongside a slight accuracy increase compared to baseline (from 62.99% to 65.03%); and DVD-B delivered a shape bias of 0.83, close to the human range, at a marginal 4.3%-point accuracy decline over baseline. Thus, DVD training provided an increase in shape-selectivity by 105.88% (DVD-P), 141.18% (DVD-B) and 164.71% (DVD-S) over our gold-standard baseline with high-res training. A category-specific depiction of shape bias for DVD-P, the ResNet-50 baseline, and human observers is provided as a supplement (Fig. S1) alongside illustrative examples showing shape/texture behaviour for DVD, ResNet-50 baseline, and ChatGPT-4o.

Importantly, the shape-bias observed in our DVD-trained ResNet-50 model not only outperforms the corresponding control models by a large margin, but also all other contemporary ANN models tested (Fig. 2c), setting the state-of-the-art in shape-selectivity and closing the gap to the human range of shape-selectivity. These included standard supervised convolutional neural networks (CNNs), vision transformers (ViTs)[52,53], self-supervised models, and multimodal vision-language models. Note that these models include DNNs trained on orders of magnitude more data, as well as models operating at orders of magnitude more model parameters, highlighting the effectiveness of DVD training, even in smaller-scale data/model scenarios.

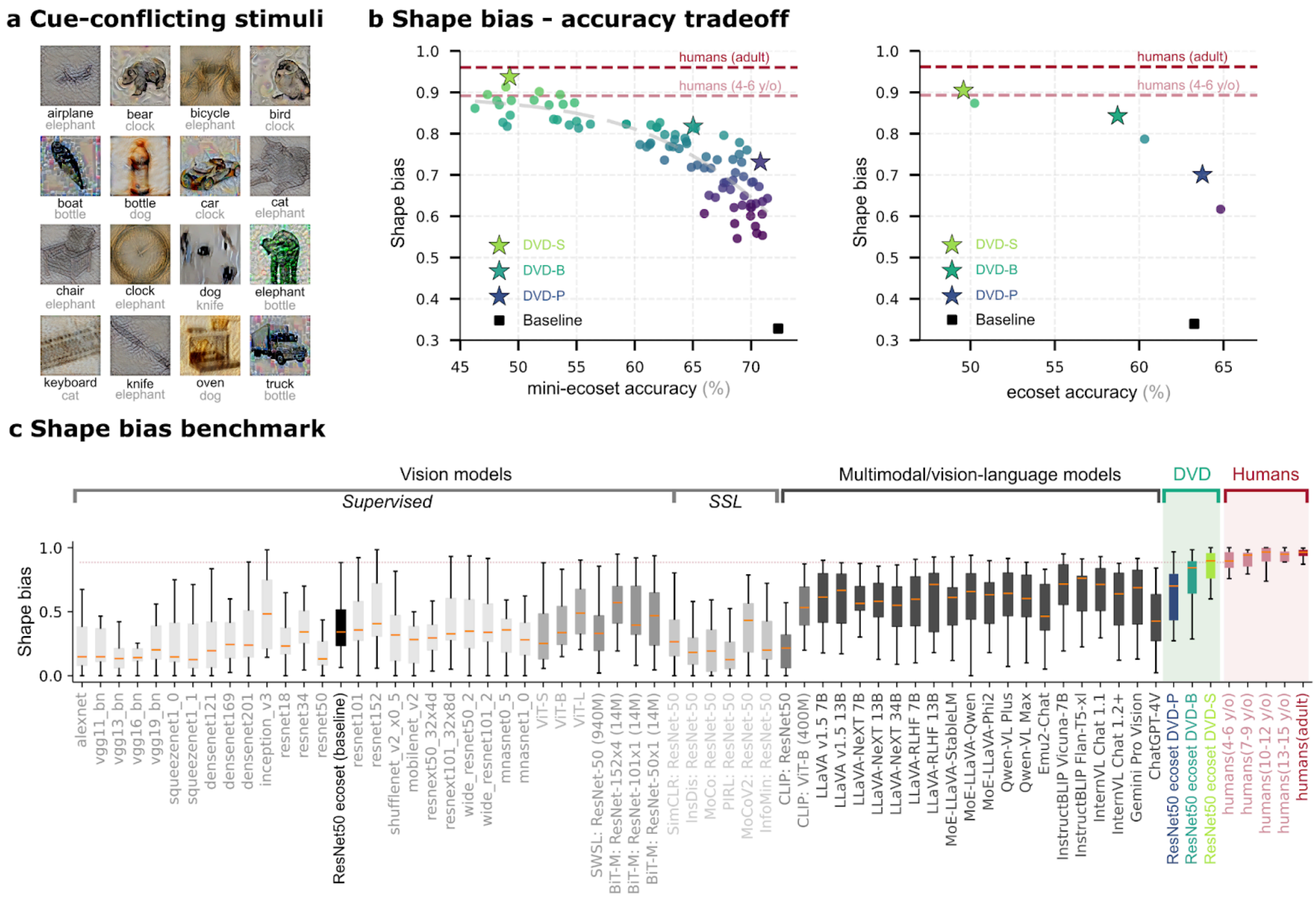

**Figure 2 | Models trained with DVD close the gap to human-level shape bias.** ***a.*** *Example cue-conflict stimuli from the 16 object categories tested. Shape and texture cues are intentionally mismatched (e.g., an object with the shape of an airplane is shown with the texture of elephant skin), to test which of the two feature types underlies human and AI decisions. Shape categories are indicated in black, texture categories in grey.* ***b.*** *Trade-off between object recognition accuracy and shape bias across models trained on mini-ecoset (0.28 million images) and full ecoset (1.5 million images). Three exemplary DVD-trained models were selected for subsequent analyses: a high recognition performance variant (DVD-P); and a balanced variant (DVD-B), which achieves close to human-level shape bias while incurring only a modest cost in accuracy; and a model prioritising shape-based decision making (DVD-S).* ***c.*** *Comparison of shape bias between DVD models, standard supervised/self-supervised vision models, large-scale multimodal vision-language models (VLMs) and humans across different age groups. Boxplots show the distribution of shape/texture biases across the 16 cue-conflict object categories (n=16; unit of analysis = category; shape bias: high values, texture bias: low values), boxplot midline represents the median, bounds indicate the 25th and 75th percentiles, and whiskers extend to the 5th and 95th percentiles. Each model corresponds to one independently trained instance (one seed). Human data: 4-6 years (n = 21), 7-9 years (n = 11), 10-12 years (n = 12), 13-15 years (n = 8) and adults (n = 97). The dotted line indicates the lower bound of average human-level shape bias (88% average shape-based decisions of 4-6-year-old humans).*

The generality of our results was ensured by a sequence of additional experiments. First, we expanded training from mini-ecoset[46], and ecoset[47] to also include ImageNet-1K[48] training. Across all three datasets, DVD consistently increased shape bias compared to the baseline, with shape bias progressively rising from DVD-P to DVD-B to DVD-S. DVD-S achieved shape bias close to human levels regardless of the dataset (Fig. 3a), and the same pattern was observed for a self-supervised DVD-B model trained with SimCLR[54] and BYOL[55] objectives on mini-ecoset, ecoset and ImageNet-1K (Fig. S16). Second, in addition to ResNet-50 training, we included 8 further contemporary ANN architectures, including various convolutional and ViT models, all trained according to the DVD-B scheme (Fig. 3b). In every case, DVD-B preprocessing substantially increased shape bias, confirming the broad effectiveness of the method in promoting shape-based representations across model types.

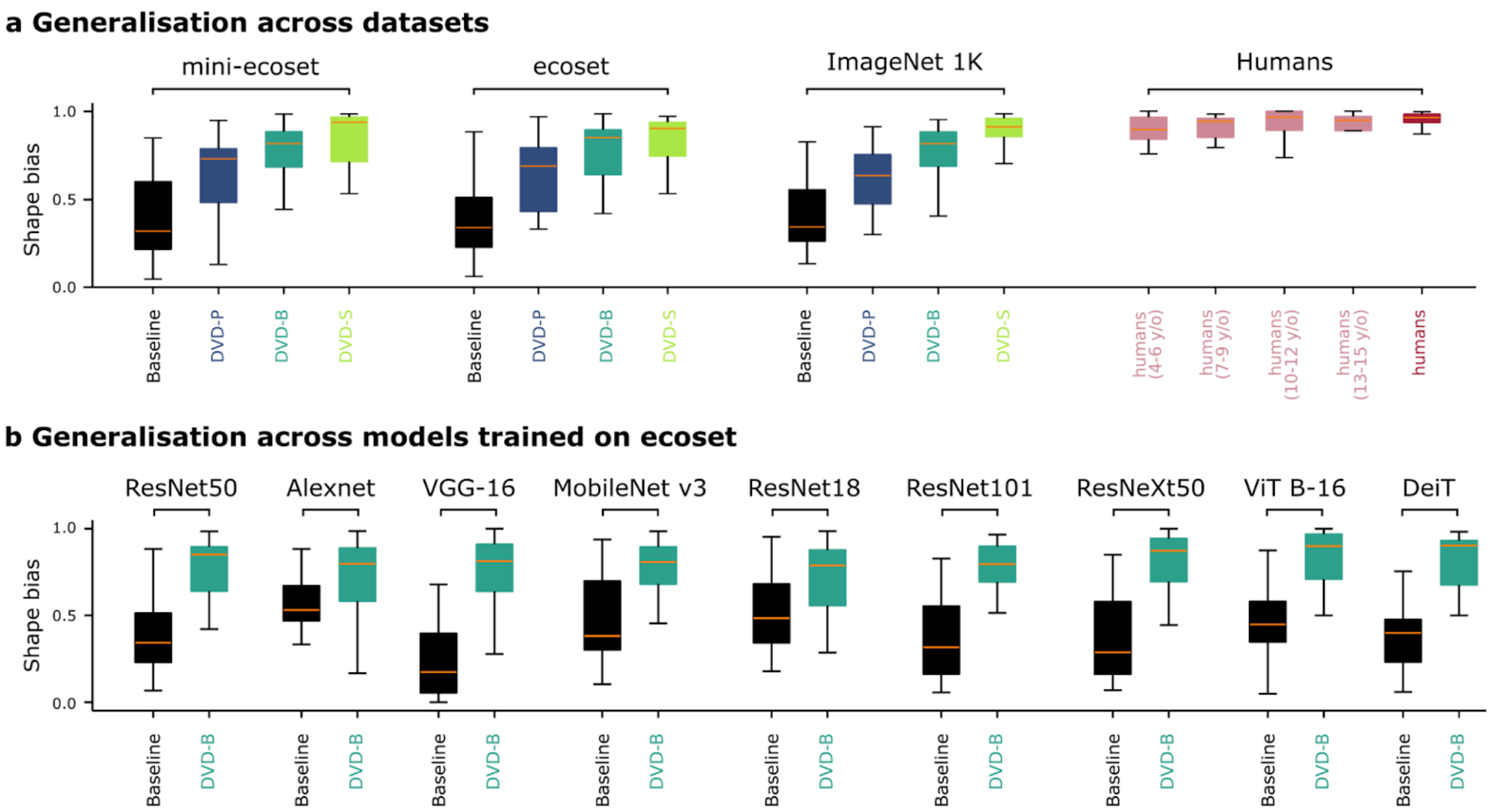

**Figure 3 | Results generalisation across vision datasets and ANN architectures.** ***a.*** *Generalisation of results across training datasets (0.28 million mini-ecoset, 1.50 million ecoset and 1.28 million ImageNet-1K). Comparable generalisation is also observed for self-supervised DVD-B models (see Fig. S16).* ***b.*** *Generalisation of results across a range of DNN architectures, including both CNNs and ViTs. Boxplots summarise category-wise shape bias across the 16 cue-conflict object categories (n = 16; unit of analysis = category) for each model (one seed; independently trained model instance). Human sample sizes and boxplot conventions are identical to Fig. 2c.*

We quantified the emergence of shape bias over time by analysing the DVD-B networks across training epochs as they progressed through the developmental trajectory (in line with training, 2 months of human development equate to 1 training epoch). We observe a strong shape bias very early in training, which remained relatively stable as learning progressed (strong shape bias at an early stage corresponding to ~20 months of age, Fig. 4a and Fig. S2; robust across 3 seeds), akin to psychophysical evidence that human infants are already shape-biased by around two years of age[56]. In contrast, the ResNet-50 baseline model remained consistently texture-biased across training (average shape bias ~0.43).

To contrast whether our observed benefits are due to the precise trajectory or solely a function of the distributions of image augmentations that DVD implements, we ran a control experiment in which the sensitivity values for acuity, contrast, and colour were randomly shuffled in time rather than presenting them in chronological sequence. As shown in Fig. S8a, the original (chronological) DVD-B consistently produced more shape-based and fewer texture-based choices than the shuffled condition, particularly during the middle and later phases of training. The difference was also reflected in the evolution of shape bias: after the first 20 months of simulated development, the chronological DVD-B maintained a markedly higher shape bias than its shuffled counterpart (Fig. S8b).

Next, we explored how DVD affects the decision-making strategies of our models - a key step in linking training regime to classification behaviour. Gradient-weighted Class Activation Mapping (Grad-CAM; see Methods), run across the cue-conflict stimuli, revealed that our DVD-trained model predominantly relied on features from patches distributed over the shape, whereas the ResNet-50 baseline model assigned greater importance to features from small patches or highly distributed background patches (Fig. 4b; see Methods for analysis details). A stronger focus on shape-related features, rather than local textures or background texture patches, indicates a perceptual strategy more aligned with human visual decision making.

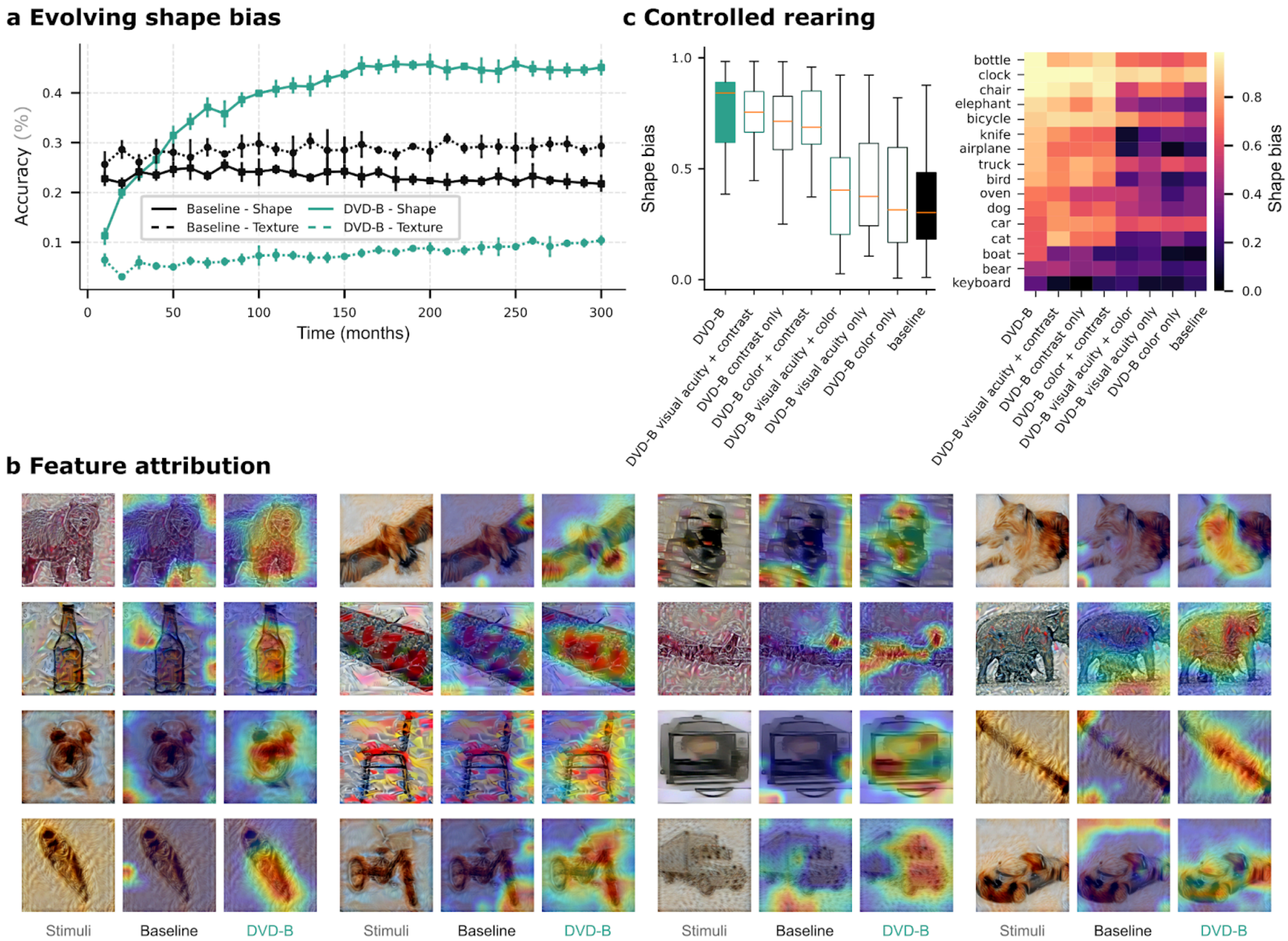

**Figure 4 | In-depth investigation of the emergence of shape bias in ANNs and the underlying feature selectivity.** ***a****. Evolving shape bias in the DVD-B model from birth to 25 years old (300 months; 2 months/epoch for DVD-B; see also Fig. S2 for the evolution of the shape bias metric), as reflected by a rising proportion of shape-based choices on cue-conflict trials (i.e., predictions matching the shape category). In contrast, the baseline model exhibits a consistent preference for texture-based choices. Mean across three independently trained seeds shown per model (n = 3); error bars indicate the ±1 standard deviation across seeds.* ***b****. Feature attribution using Grad-CAM, indicating that DVD-B relies more strongly on image regions corresponding to the object in question (data shown for a model instance; see also Fig. S4 for feature importance visualisation using layer-wise relevance*

*propagation). **c**. Controlled rearing experiments to study the role of different visual aspects in development (based on DVD-B) indicate a strong impact of contrast sensitivity development (see Fig. S11 for an illustration of its critical role). Including all 3 aspects of visual development yields the most consistent shape bias across categories. Boxplots summarise category-wise shape bias across the 16 cue-conflict object categories (n = 16; unit of analysis = category); data of each rearing condition based on a single instance (one seed; independently trained model instance). Boxplot conventions as in Fig. 2c.*

To understand the impact of each of the three developmental aspects, acuity, contrast, and colour, we performed controlled rearing experiments, which systematically tested all seven combinations for their impact on shape bias. This systematic manipulation revealed, in contrast to the previous focus on blur in the literature, a central role of contrast sensitivity development as a key driver of shape bias in our models. Models trained with contrast sensitivity development achieved a shape bias of 0.73 (compared to 0.85 observed for the original DVD-B model (Fig. 4c, Fig. S3). In contrast, models exposed only to visual acuity development, or to both visual acuity and chromatic sensitivity development, exhibited far smaller increases in shape bias (0.41 and 0.44, respectively), relative to the baseline shape bias of 0.34. To examine why contrast sensitivity exerts such a strong influence, we analysed its role under low spatial-frequency conditions (corresponding to low visual acuity; Fig. S11). This illustrative approach demonstrates that high-contrast signals primarily preserve global object structure, whereas low-contrast counterparts emphasise textures with little shape coherence. This analysis thereby highlights contrast sensitivity as an important source of shape information, going beyond merely considering low-frequency signals (i.e. visual acuity), consistent with its dominant contribution to the emergence of shape bias.

Together, our results indicate that taking inspiration from the human developmental visual diet is a highly effective way to bias AI vision systems towards more human-like perceptual decision making, delivering the state-of-the-art in shape bias across a large range of open-source and closed-source models, including AI foundation models.

### *DVD-trained models exhibit higher recall for abstract shapes embedded in complex contexts*

Additional evidence for the misalignment between human and AI vision comes from experiments on abstract shapes[7]. Humans effortlessly recognise abstract shapes, i.e. recognizable forms defined by their overall configuration rather than specific textures or colours, even when embedded within complex natural scenes (Fig. 5a). In contrast, current AI systems, including large-scale vision-language models (VLMs), struggle to detect such shapes, and instead rely on scene-based cues[7].

To assess the efficacy of DVD in enhancing abstract shape recognition within naturalistic contexts, we evaluated DVD-trained models alongside gold-standard baseline models and a large array of DNN architectures using the IllusionBench dataset - a benchmark that specifically tests shape recognition through scene-embedded visual arrangements (6874 images, 16 shapes, 11 background scenes, see Methods). In this benchmark, shape recall reflects a model's tendency to make predictions based on abstract shape cues, while scene recall reflects its tendency to rely on background scene or texture cues. Conventional CNN models, such the ResNet-50 baseline, achieved a shape recall of 8.71%. The best-performing Vision Transformer (ViT-L) reached 17.13%. Large-scale VLMs, including CLIP-RN50 and CLIP-ViT B-32, attained 16.72% and 15.45%, respectively, and ResNet-50 trained on a style-transferred dataset reached 14.32%. In contrast, our DVD-trained ResNet-50 (DVD-S, trained on ImageNet-1K) significantly outperformed all other tested models by achieving a shape recall rate of 36.21% (Fig. 5b). Notably, large AI foundation models, even though explicitly prompted to spot the shape in the background, were outperformed (shape recall for Llama-4-Scout (12.47%), Gemini-2.0-flash (21.24%), and ChatGPT-4o (15.17%)). This again demonstrates the ability of DVD training to achieve more human-aligned representations even when operating at orders of magnitude less data and model parameters. DVD-S models also displayed substantially lower reliance on scene cues (20.07% scene recall) compared to approximately 60% scene recall rates in other models. Shape recall improved systematically across the DVD family as the strength of early visual developmental constraints increased, from 16.43% in DVD-P to 36.21% in DVD-S, while texture-based recall declined (from 46.32% to 20.07%). Further analyses of the internal representations of the various models revealed a striking pattern that differentiates DVD-trained models from other models. Visualising model embeddings of IllusionBench stimuli using t-SNE[57] (Fig. 5c), we observe that only DVD-trained models distinctly clustered images by abstract shape category, whereas ResNet-50 baseline, CLIP-ViT B-32, as well as ResNet-50 models trained on style-transferred ImageNet predominantly clustered images based on scene context. The importance of this extension becomes apparent when considering that ResNet-50 models trained on style-transferred ImageNet, despite being among the most shape-biased according to the Geirhos metric[1], still predominantly clustered images by scene context, indicating that such models still lack a genuine configural processing. Together, these analyses demonstrate that DVD training substantially alters the visual features learned to perform the task, promoting the spatial integration of distributed shape cues.

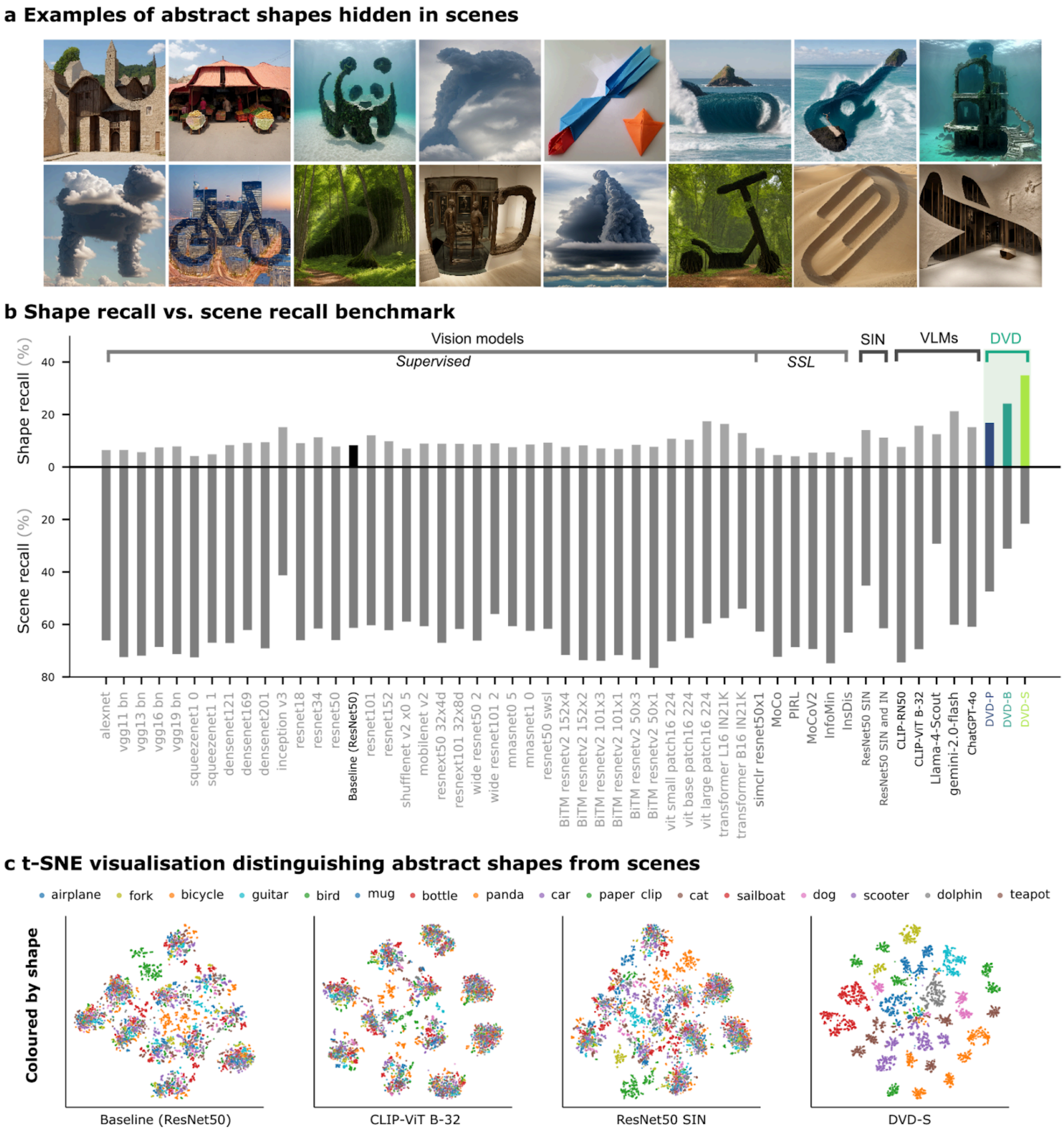

**Figure 5 | Models trained with DVD exhibit enhanced recognition of abstract shapes hidden within complex scenes. a.** *Example IllusionBench stimuli of abstract shapes hidden in scene contexts.* ***b.*** *Shape recall vs. scene recall benchmark for DVD-trained models, as well as a large range of control models, including AI foundation models and large-scale multimodal vision-language models.* ***c.*** *t-SNE embeddings of IllusionBench images, colour coded by object category. Only DVD-trained models cluster images according to their abstract shape categories, whereas other models predominantly cluster images based on their scene contexts (also see Fig. S5).*

***DVD training enhances robustness to various types of image degradation***

Having established the reliance on spatially-integrative shape-like decisions in DVD-trained DNNs, we tested the models for their behavioural robustness in the face of degraded input[4] - a hallmark of our robust human vision. In this context, we treat robustness as a probe of whether DVD's shift from predominantly texture-based to genuinely shape-based visual processing yields emergent benefits under challenging viewing conditions. Cases of image blur, image noise, quality deficits, as well as environmental factors such as rain and snow, were considered. While human performance is highly robust, and degrades gracefully across severity levels, prior tests of AI vision models revealed a dramatic drop in performance even at low levels of image degradation.

To examine whether DVD is able to close the gap between biological and artificial vision, we first compared a trained DVD-B model with a ResNet-50 baseline under increasing levels of Gaussian blur applied to images depicting objects and faces (see Methods). Blur is a particularly challenging degradation because it directly disrupts the fine spatial details and edges that many artificial vision systems rely on for recognition, while human recognition abilities degrade gracefully[14]. This makes blur an important test case for evaluating whether a model has learned to rely, akin to humans, on global, integrative information rather than relying on local detail. As blur severity increases, the baseline model shows a sharp drop in recognition accuracy, while the DVD-trained model maintains significantly higher accuracy, approximately following human behavioural data (Fig. 6a; human data from Jang et al., 2021[14]).

We then evaluated baseline and DVD-trained models across 16 additional types of degradations (4 types of noise, 4 types of blur, 4 types of weather disruption, 4 types of image quality deficits, Fig. 6b; see Methods), each at 5 severity levels from mild (1) to strong (5). At high severities, DVD accuracy was 2 times higher than baseline models for noise, blur, and challenging weather effects, and 3-4 times higher for image quality deficits. These results demonstrate that DVD training greatly improves robustness under challenging conditions (this also holds for DVD-S and DVD-P, see Fig. S13).

***DVD training enhances robustness to white- and black-box adversarial attacks***

Adversarial attacks, i.e. small, but targeted image perturbations, can dramatically bias and impair model performance[5,6,9]. While often imperceivable for humans, these attacks pose yet another major challenge for AI vision systems and showcase the problems in terms of behavioural alignment. We tested whether DVD training improves adversarial robustness, again compared to control models trained with the same architecture, training-dataset and task, but following the gold-standard of high-res training. DVD-trained models exhibit marked improvements under both black-box[58,59] as well as white-box[60] adversarial attacks (Fig. 6c). From the family of black-box attacks, a DVD-trained model achieves about 50% higher accuracy for L2-based Gaussian and uniform noise at severity 5, and about 3 times higher accuracy under impulse noise at the same severity. Against white-box attacks, DVD improves performance against Fast Gradient Sign Method (FGSM[5]; from 17% to 40%), against the Fast Gradient Method (FGM[5]; from 13% to 32%), and against a Projected Gradient Descent attack (PGD[9]; from 11% to 39%) again at severity 5. These findings confirm that DVD training renders models substantially more robust against adversarial attacks.

To better disentangle the impact of individual developmental factors on robustness, we considered the models from the controlled rearing analysis that were trained with acuity, contrast, or colour maturation alone (Fig. S10). Mirroring the shape bias results, contrast sensitivity emerged as the dominant contributor, consistently enhancing robustness across most naturalistic corruptions and adversarial perturbations. Acuity and colour provided selective benefits under blur-based corruptions (especially motion, defocus) and JPEG compression but were largely ineffective against noise and weather-based distortions. Combining all three factors produced the strongest and most generalizable gains in robustness, indicating that the integrated developmental trajectory affords advantages beyond those of any single factor alone.

Finally, to benchmark the strength of DVD training, we compared it against adversarial training (Fig. S9). Specifically, we trained control models with PGD-based adversarial training (AT; $\varepsilon$ = 1e-3, 1/255, and 4/255[61,62]). While such a model performed well against related white-box attacks (FGM, FGSM), their generalisation to black-box attacks was limited, falling below that of DVD-trained models. For example, DVD-B achieves about twice the accuracy of adversarially trained controls under impulse noise at severity 5. Moreover, DVD training required 4.62 times less computation time than AT (Fig. S15a) and typically outperformed AT across a broad range of naturalistic degradations (Fig. S9). Taken together, these comparisons suggest that DVD training provides a competitive alternative to AT. Additionally, DVD confers benefits not attained through AT alone, including close to human-level shape bias, shape-based feature selectivity in attribution maps, and state-of-the-art abstract shape recognition performance (see Fig. S15b-d).

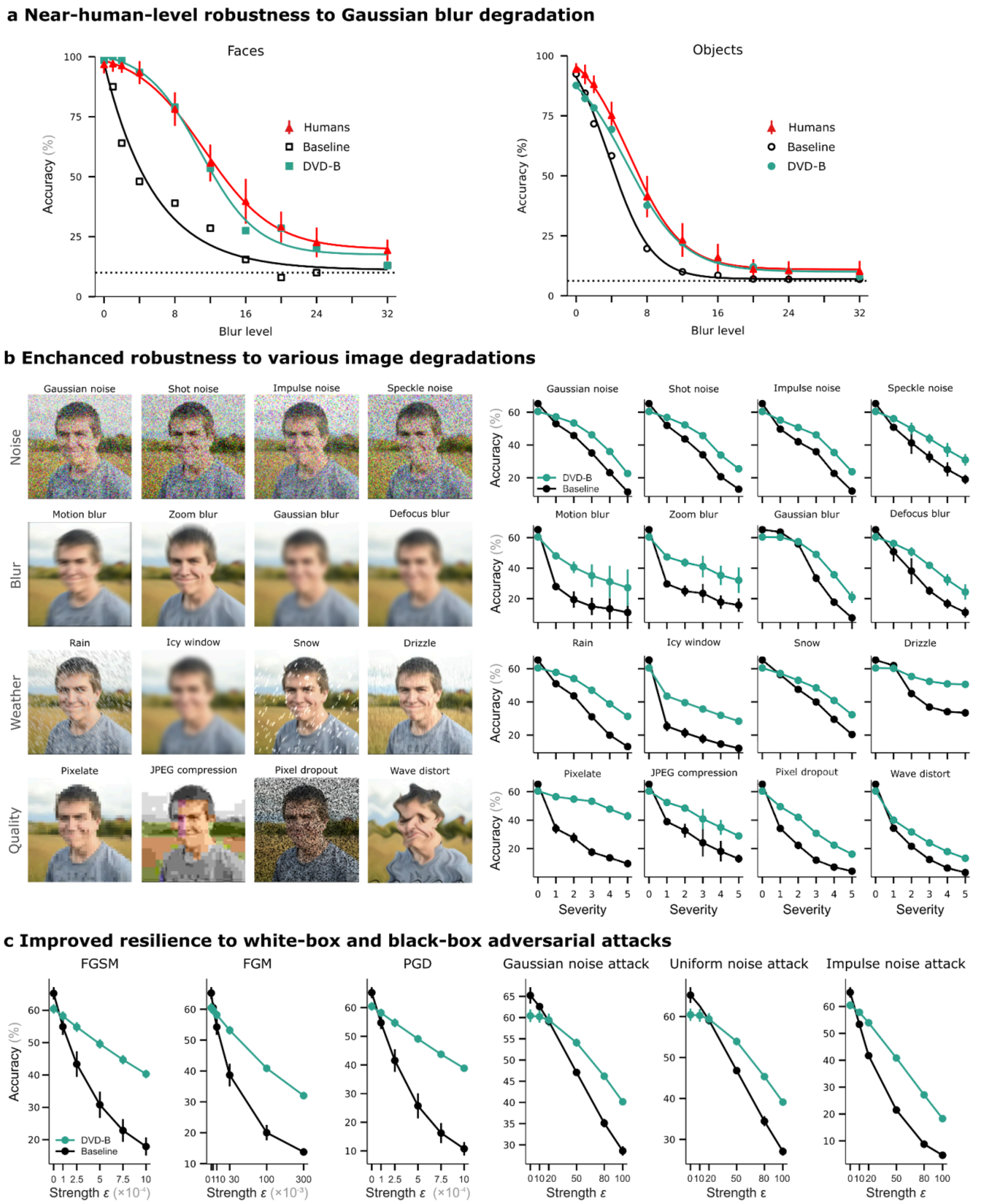

**Figure 6 | ANNs trained with DVD exhibit enhanced degradation robustness to image degradations and adversarial attacks.** ***a.*** *Alignment of DVD-B-trained and baseline models with human response profiles under increasing severity of blurring for both face- and object recognition (humans: n = 20 participants, mean across participants shown; error bars indicate ±1 standard deviation across participants; models: single instance, independently trained).* ***b.*** *Resilience to various image degradations, including different noises, blurring, weather, and image quality deficits for DVD-B-trained and baseline models (mean across seeds shown, n = 3; error bars indicate ±1 standard deviation across seeds; see Fig. S9 for comparison with an adversarially trained benchmark model, and Fig. S10 for comparisons across models trained solely with one of the three individual visual*

*factors).* ***c.*** *Robustness estimates of the DVD-B-trained model versus baseline under adversarial perturbations (mean across seeds shown, n = 3; error bars indicate ±1 standard deviation across seeds). White-box gradient-based attacks: Fast Gradient Sign Method (FGSM), Fast Gradient Method (FGM), and Projected Gradient Descent (PGD). Black-box attacks:* $L_2$*-norm additive Gaussian noise,* $L_2$*-norm additive uniform noise, and impulse noise (see also Figs. S9, S10).*

## Discussion

Acquiring human-like shape-based, robust vision has been a long-standing challenge for AI. Instead of "scaling up" data and architecture size, we here demonstrate that inspiration can be found in biology. Across a range of experiments with different datasets and architectures, we show that exposure to the visual developmental trajectory of humans, from newborn to 25 years of age, yields significantly more shape-based and robust, human-aligned vision systems. DVD-trained models are state-of-the-art in shape-based decision making, outperforming large-scale AI foundation models[63] and closing the gap to human behaviour. The same models also exhibited the ability to recognise abstract shapes hidden in complex backgrounds — again a test that large-scale foundation models fail[7]. In addition, DVD-training closely aligned our models with human psychophysical data upon Gaussian image degradations[14] and rendered them more robust to all other image perturbations tested[4,15]. Finally, DVD-trained models were shown to be more robust to a variety of black- and white-box adversarial attacks. Thus, DVD provides a general training framework for achieving shape-based visual processing models that get close to human-level shape-based, state-of-the-art abstract shape recognition performance, with increased adversarial robustness and degradation robustness emerging as natural consequences. Together, our analyses demonstrate that DVD-training yields models with a fundamentally different, robust feature set, upon which they base their decisions. Rather than relying on distributed local features, such as textures, they derive their robustness from operating on integrative shape-based features.

By running controlled rearing experiments, we find that the development of contrast sensitivity, rather than visual acuity or colour, played a pivotal role in obtaining shape-selective and robust performance. This runs counter to the intuitions behind previous studies that mainly emphasised reduced visual acuity, simulated through image blurring[14–19], as a key driver of shape bias. While visual acuity sets the upper bound of resolvable spatial frequencies, contrast sensitivity determines the minimal amplitude that can be detected at a given spatial frequency. Crucially, not all low-frequency content supports shape: weak amplitudes convey little beyond uniform texture, whereas low-frequency content with high amplitude primarily encodes global structure. Thus, contrast sensitivity constitutes a critical, and previously less appreciated, determinant of shape-based processing. In addition, contrast sensitivity matures more slowly than visual acuity; its prolonged immaturity may thus exert a deeper and more lasting influence on the development of shape perception. Note, however, the natural complexity of human contrast sensitivity across spatial frequencies during development is not fully captured by frequency-domain operations implemented via hard thresholding, contrast sensitivity function multiplication, or even with one-octave frequency pooling. Future work may investigate the effects of a closer alignment with human contrast sensitivity development. Also, absolute contrast-threshold measurements differ across developmental paradigms and may reflect non-visual factors; therefore, our implementation is intended as an approximate abstraction rather than a direct quantitative mapping of threshold contrasts across ages.

Random shuffling experiments further demonstrated that respecting developmental order confers a distinct advantage: shape-biased behaviour emerges stronger under chronological schedules than under randomly-shuffled exposure, highlighting the benefit over random data augmentation schemes. In the case of humans, for whom training and testing cannot be strictly separated, an early phase of "poor" vision proves advantageous, compared to continuously, and randomly interleaved sensory degradation, as implied by augmentation training, due to our prolonged phase of parental care[12].

While the current work takes inspiration from visual development, many promising aspects of biological vision remain to be explored for their implications on AI vision. Among others, these include overt visual information sampling[64,65], topographic ordering of visual information[66,67], the differential roles of magno- and parvocellular pathways[19], the co-development of different sensory modalities[68], the bodily action repertoire[69,70], and language abilities[71,72]. A complementary direction in modelling the development of vision involves using egocentric videos to map the changes in the distribution of objects and their views[73], wherein the curriculum has also been shown to be important in developing rich visual features[74]. Joint modelling of both the distribution of exposure to objects and their views, through egocentric videos, and the distribution of image features, through our DVD transformations, would further close the gap between the developmental diet of humans and AI systems. Future work may explore the impact of this combination on shape-based, robust AI vision.

The current work evaluated AI vision models across a large battery of tests. This variety proved important, as previous solutions to some tests were found to fail at others. For example, in evaluating shape-based object recognition, style-transferred shape-texture cue-conflict images have been central to the AI community. However, our results indicate that merely scoring high on that shape bias benchmark does not necessarily correspond to shape-based visual processing: models trained with style-transferred images showed high shape bias but poorly detected abstract shapes embedded in natural scenes. This finding underlines the importance of testing for shape-based processing via multiple benchmarks designed to tease apart various properties of shape information[7,75].

In addition to the benefits for machine vision, the current approach may have implications for our understanding of how human vision may come to be so robust, indicating that "starting out poor" is an efficient way for the brain to learn highly robust visual features[10–12]. As our experiments demonstrate, an early reliance on configural shape cues, enforced by missing high spatial frequency as well as overall frequencies with low power, may establish a perceptual bias that persists throughout human development, guiding human vision towards a shape-dominant visual processing strategy. Going forward, our models may serve as an important test-bed for deriving predictions on human visual developmental features, and provide a valuable starting point for explorations of the effects of medical interventions, such as the removal of congenital cataracts[76] and ways to mitigate the residual downsides of this intervention.

## Methods

### Neural network architectures and training

*Developmental visual diet (DVD) preprocessing pipeline*

DVD is implemented as a data preprocessing pipeline (Fig. 1a) that emulates the maturation of 3 visual aspects: visual acuity[20–34], chromatic sensitivity[35–38], and contrast sensitivity[39–45]. Quantitative psychophysical data from birth to 25 years (0-300 months) were collated and fitted with smooth, monotonic functions (Fig. 1b, Fig. 1c, Fig. 1d); the resulting curves drive a data transformation schedule that spans the full training epochs. Note that both retinal immaturities and cortical development contribute to early visual limitations; here we rely on psychophysical data that capture the combined influence of these factors. This design enables a preprocessing pipeline that can be readily integrated into the training of diverse network architectures. Disentangling retinal and cortical contributions remains an important direction for future work.

*Visual-Acuity simulation via Gaussian blurring*

Visual acuity is modelled by applying a Gaussian blur (σ) to images, in accordance with measurable changes in Snellen acuity (Fig. 1b). Snellen acuity across ages was acquired through psychophysics studies [20–34], and the data are fit with a double-exponential function. Following prior work on visual acuity simulation[18], a 15° visual angle object viewed at 60 cm allows a 20/20 observer to resolve ≈ 450 cycles, while a 20/600 observer can resolve only ≈ 15 cycles. For an image width of $w$ pixels, we compute $\sigma$ such that all spatial frequencies above 1/(w/15) are attenuated. Convolution with this filter produces an image approximating 20/600 acuity when σ=4 for an image size of 100×100[18], which allows for converting any other Snellen fraction $x$ into Gaussian blur $\sigma$ :

$$\sigma(w,x) \; = \frac{4 \cdot (20/600) \cdot w}{100 \cdot x}$$

*where $w$ is the image width in pixels, and $x$ is the Snellen visual acuity;* increasing σ simulates reducing *Snellen* visual acuity.

*Contrast sensitivity simulation via frequency-domain thresholding*

Contrast sensitivity is modelled by applying an age-dependent amplitude threshold to each image's frequency-domain representation, selectively removing weak signal components that would be imperceptible to a developing human visual system. At each training epoch, the model is assigned an age *t* in months, and the corresponding image is processed to reflect contrast sensitivity limitations appropriate for that age. Specifically, we compute the image's discrete Fourier transform and extract the power spectrum per RGB channel. A dynamic amplitude threshold $T_t$ was defined to suppress frequency components whose power falls below the model's sensitivity at that age. This threshold is defined as:

$$T_t \; = \frac{P_{\max}\, \beta\, \bigl(1 - C_t\bigr)}{\max\bigl(\lfloor \frac{t}{\lambda} \rfloor\, ,\, 1\bigr)}$$

where $P_{\max}$ is the maximum spectral power across all RGB channels for the given image, $C_t$ is the empirically estimated contrast sensitivity at age t normalized to the range [0, 1], derived from quantitative human psychophysical data[39–44] tracking the maximum contrast sensitivity, across spatial frequencies, from birth to early adulthood (0-300 months; see Fig. 1c; function fitted according to Dekker et al.[40]). $\beta$ and $\lambda$ are fixed hyperparameters. $\beta$ determines the amplitude threshold in the frequency domain corresponding to the contrast sensitivity at birth, while $\lambda$ controls the mapping between the rate of threshold decay and the corresponding increase in contrast sensitivity over time.

Frequencies whose amplitude falls below $T_t$ are set to zero, effectively discarding image information that would be undetectable at that stage of development. By default, this procedure implements a strict cutoff, nulling all subthreshold signals. To approximate the gradual nature of human perceptual thresholds, we also tested sigmoid thresholding (Fig. S7a) to provide a smoother transition between visible and invisible contrasts, which yielded qualitatively consistent results (Fig. S7b). Furthermore, we implemented a CSF-multiplication method that modulates all spatial-frequency components rather than applying a hard thresholding cut-off, and we also introduced frequency pooling over one octave in the frequency domain; together, these implementations reduced visible preprocessing artifacts (Fig. S14a). Models trained with CSF multiplication, with or without frequency pooling, yielded results that were consistent with those obtained using the thresholding-based approach (Fig. S14b-d). The resulting contrast-limited image is reconstructed by inverse Fourier transforming the filtered spectrum, yielding a naturalistic simulation of age-dependent visual experience for model training.

In the main analyses, we modelled developmental vision using a single measure of peak contrast sensitivity across ages as a concise preprocessing step for AI training (Fig. 1b). However, infant contrast sensitivity varies not only with age but also across spatial frequencies (see Fig. S6a-b). To capture these dynamics more faithfully, we implemented a developmental contrast sensitivity function that evolves over time, reflecting the gradual maturation of sensitivity across frequencies, spanning 1 to 32 cycles per degree and newborn to 25 years of age (Fig. S6c). Models trained with this full CSF-based transformation produced outcomes that were largely consistent with those obtained using the simplified peak-sensitivity approach (Fig. S6d).

*Chromatic sensitivity simulation via colour fidelity Interpolation*

To model age-dependent changes in chromatic sensitivity, we applied a pixel-wise linear interpolation between grayscale and full-colour images, modulated by empirically derived chromatic sensitivity values ranging from 0% (newborn) to 100% (adult). These values represent the relative ability to perceive colour differences compared to mature observers. At each training epoch corresponding to developmental age $t$ (in months), we define a chromatic sensitivity factor $S_t$ in [0, 1], fitted from behavioural data[35–38] (see Fig. 1d; function fitting according to Knoblauch et al.[38]). This parameter governs the image transformation applied during training:

$$I_t = (1 - S_t) \cdot I_{\text{gray}} + S_t \cdot I_{\text{RGB}}$$

where $I_{\text{RGB}}$ is the original full-colour image, $I_{\text{gray}}$ is its grayscale (luminance-only) counterpart, and $I_t$ is the resulting developmentally adjusted image at age t. This continuous schedule allows the model to experience a progressive increase in colour fidelity, mirroring the maturation of chromatic sensitivity over early development and into adulthood.

*Hyperparameter tuning of DVD*

The DVD preprocessing pipeline employs three key hyperparameters to control early visual experiences during model training: $\alpha$ (months per epoch), $\beta$ (initial contrast threshold), and $\lambda$ (contrast sensitivity mapping factor). For hyperparameter tuning, we swept $\alpha$ in [1, 2, 4, 8], $\beta$ in [5e-5, 1e-4, 2e-4, 4e-4], and $\lambda$ in [50, 100, 150]. The hyperparameter sweep was performed only on mini-ecoset to characterise the accuracy–bias trade-off. For ecoset, ImageNet, across architectures (CNNs and ViTs), and from supervised to self-supervised learning, we directly applied the representative settings from mini-ecoset without further tuning (see Table S1 for the exact parameter values used in the main DVD variants).

The parameter $\alpha$ defines the temporal resolution of the training process by mapping each training epoch to a specific number of developmental months. A smaller $\alpha$ results in finer temporal granularity, allowing the model to experience more gradual changes in visual input. A larger $\alpha$ accelerates progression through developmental stages. For the main DVD variants, DVD-S ($\alpha$ = 1 ), DVD-B ($\alpha$ = 2) ,and DVD-P ($\alpha$ =4), $\alpha$ serves as a flexible proxy for early visual exposure, translating developmental months into training epochs without assuming absolute values for infant wake time, stimulus exposure, or repetition. Thus, decreasing or increasing $\alpha$ effectively controls the quantity and strength of early immature experience, enabling systematic investigation of its influence on visual representations and behaviours.

The parameter $\beta$ sets the baseline for initial contrast sensitivity by defining the starting amplitude threshold in the frequency domain, simulating newborn vision. As training progresses, this threshold is lowered to reflect improving sensitivity. The parameter $\lambda$ controls the rate at which the frequency-domain threshold decays over time, translating contrast sensitivity improvements into frequency-domain threshold attenuation as development progresses.

*Neural network architectures training procedure*

We based our main experiments on the ResNet-50 architecture as the default backbone for the baseline model and the DVD-trained models. The baseline models received conventional high-fidelity data preprocessing, whereas DVD-trained models were fed inputs transformed by the developmental visual-diet pipeline described below. We also trained a family of convolutional neural networks (CNNs) and vision transformer (ViTs) backbones (ResNet-50 by default; AlexNet, VGG-16, MobileNet-V2, ResNet-18, ResNet-101, ResNeXt-50, ViT-B/16 and DeiT-B/16[51–53] for generalisation tests across architecture; Fig. 3b). All models were trained for 300 epochs, parallelized across two NVIDIA H100 GPUs using stochastic gradient descent (momentum = 0.9, weight decay = 1e-6, learning rate = 1e-4), and received identical data augmentations including random horizontal flip, rotation, grayscale, brightness, Gaussian blur, sharpness, equalization and perspective. We avoided aggressive small crops during training, as excessively reduced input fields often omit global object structure and are therefore likely to hinder the emergence of shape-selective representations. To assess potential interactions, we also trained control models without Gaussian blur augmentation, which produced comparable shape bias outcomes (Fig. S12).

*Training datasets*

We evaluated models trained with and without the DVD using three image datasets: mini-ecoset[46] (0.28 million images, 112 categories, including 12 added to match cue-conflict categories), ecoset[47] (1.5 million images from 565 basic-level categories), and ImageNet-1K[48] (~1.3 million images, 1,000 classes).

**Model analyses**

*Shape vs texture bias*

Shape bias is measured using cue-conflict stimuli, where texture and shape cues are in conflict[1]. In accordance with Geirhos et al., we computed the shape-bias as:

$$B_{\text{shape}} = \frac{N_{\text{shape}}}{N_{\text{shape}} + N_{\text{texture}}}$$

where $N_{\text{shape}}$ is the count of cue-conflict images correctly classified according to the shape cue, $N_{\text{texture}}$ is the number classified according to the texture cue, and $B_{\text{shape}}$ denotes the shape bias. This metric captures the proportion of shape-driven decisions when faced with conflicting shape-texture visual information. Unless otherwise specified, overall shape bias is reported as the median of category-level shape biases.

*Temporal dynamics of shape bias development*

To track the emergence of shape bias throughout training, we saved weight checkpoints of both DVD-trained and baseline models at 5-epoch intervals. For each checkpoint, we computed shape bias and the corresponding shape/texture classification accuracy. Using the developmental clock parameter $\alpha$ (months per epoch), we can also translate these epochs into equivalent human developmental ages with the same visual acuity, contrast sensitivity, and chromatic sensitivity.

*Controlled rearing studies*

To disentangle the contributions of individual visual-development components, we conducted controlled-rearing experiments by training models with all possible combinations of three factors: visual acuity, contrast sensitivity, and chromatic sensitivity, resulting in seven distinct conditions: (1) all three present, (2) acuity + chromatic, (3) acuity + contrast, (4) contrast + chromatic, (5) acuity only, (6) chromatic only, and (7) contrast only. This systematic factorial design revealed that contrast sensitivity alone was sufficient to induce a significant shape bias, matching the performance observed in models trained with the full DVD pipeline.

*Feature attribution*

We applied Gradient-weighted Class Activation Mapping (Grad-CAM)[77], which generates class-discriminative heatmaps by combining the gradients of a target class with the activations of the last convolutional layer. Grad-CAM highlights the spatial regions most influential for the prediction, thereby providing a complementary, coarse-grained visualisation of the model's decision process (Fig. 4b).

In addition, we applied Layer-Wise Relevance Propagation (LRP) to trace model predictions back through the network layers[78], assigning relevance scores to input pixels based on their contribution to the final class decision. This allowed us to visualise whether models relied predominantly on shape-related or texture-related features when classifying cue-conflict images (Fig. S4).

*Image degradation robustness*

We evaluated robustness to image degradation by measuring top-1 classification accuracy on objects and faces across 10 severity levels, and on 16 types of degradations, each tested at 5 severity levels. The degradations included noise (Gaussian noise, shot noise, impulse noise, speckle noise), blur (motion blur, zoom blur, Gaussian blur, defocus blur), weather effects (rain, icy window, drizzle, snow), and quality impairments (pixelate, JPEG compression, pixel dropout, wave distortion). Accuracy was recorded at each severity level to quantify the model's resilience under progressively challenging conditions across all degradation types.

*Adversarial robustness*

Adversarial robustness was quantified by measuring top-1 accuracy under both black-box[58] and white-box attacks[60] at increasing intensity levels. Black-box attacks included: $L_2$-based Gaussian noise, uniform noise, and salt-and-pepper noise, each applied at perturbation amplitudes of 10, 20, 50, 80 and 100. White-box attacks comprised FGSM[5] ($\varepsilon$ = 1e-4, 2.5e-4, 5e-4, 7.5e-4, 1e-3), FGM[5] ($\varepsilon$=1e-3, 1e-2,3e-2, 1e-1, 3e-1), and PGD[9] ($\varepsilon$=1e-4, 2.5e-4, 5e-4, 7.5e-4, 1e-3). Accuracy was recorded at each level to assess model stability under increasingly strong adversarial perturbations.

*Abstract shape recognition*

We evaluated abstract shape recognition using the IllusionBench dataset[7], in which 16 distinct shape categories (airplane, bicycle, bird, bottle, car, cat, dog, dolphin, fork, guitar, mug, panda, paper clip, sailboat, scooter, and teapot) are subtly embedded within 11 naturalistic scenes: bazaar market, city, medieval village, museum, Times Square, underwater ruins, cloud, forest, ocean, origami, and sand dune.

This benchmark tests a model's capability to detect global shape configurations embedded in natural scenes. We evaluate all models, vision-language models (VLMs) and standard vision models, using ImageNet-1K predictions mapped to 16 shape super-classes and 11 scene super-classes (same across models). Shape recall measures how often the predicted class belongs to the correct shape super-class. Scene recall measures how often the predicted class belongs to the correct scene super-class. For each dataset, recall is reported as the percentage of images where the correct shape or scene super-class was identified.

*Evaluation setup for multimodal large language models*

We evaluated multimodal large language models(LLMs) / vision–language models (VLMs) (Llama-4-Scout, Gemini-2.0-Flash, and GPT-4o ) with the same constrained instruction:"You are an ImageNet classifier. Choose exactly one class label from the list provided by the user. Respond with only that class name—no quotes, no punctuation, no additional words. If uncertain, guess the most likely class. Never apologise or mention uncertainty. Classify this image among the 1000 ImageNet classes: \n{classes}\n". We adopted a closed-set classification prompting scheme, similar to those employed in previous VLMs evaluations (Zhang et al., NeurIPS 2024[79]; Gavrikov et al., ICLR 2025[63])

For IllusionBench-IN, images were resized to 224 × 224 pixels and presented as single-image requests, with exactly one generation per image and no retries (across all 6874 images). Model-specific decoding settings were as follows. Llama-4-Scout used deterministic greedy decoding (do_sample = False, num_beams = 1, max_new_tokens = 32), with no temperature specified. Gemini-2.0-Flash was run with API default parameters, including a sampling temperature of 1.0. GPT-4o was likewise evaluated under API defaults, with no explicit temperature or top-p setting. For cue-conflicting stimuli analyses, we obtained the results of multimodal LLMs through Gavrikov et al[63].

*t-SNE embedding visualisation*

To inspect the internal representations induced by DVD training, we extracted feature activations from the penultimate layer of each model (the last average-pooling layer for CNNs or the final pre-logit outputs for transformers) in response to the full IllusionBench dataset. These high-dimensional activation vectors were reduced to 50 principal components via PCA, a standard preprocessing step that stabilises t-SNE[57] and mitigates noise. The reduced representations were then embedded into two dimensions using t-SNE with a perplexity of 30, PCA-based initialisation, and 1,000 iterations (random seed = 42). The resulting 2D embeddings provide a clear visual summary of how models cluster stimuli based on learned feature similarity (Fig. 5c and Fig. S5).

## Data availability

All analyses of human and model data were performed in custom Python software, making use of PyTorch, numpy, and scikit-learn packages, among others. The code required to reproduce our results is available via GitHub at https://github.com/KietzmannLab/DVD under an MIT license and a citable DOI (10.5281/zenodo.17192363).

## Code availability

The code required to reproduce our results can be found on GitHub at https://github.com/KietzmannLab/DVD.

## Acknowledgements

The authors acknowledge support by the European Research Council (ERC) StG grant 101039524 TIME (Kietzmann), ERC Consolidator grant ERC-CoG-2024101123101 (Cichy), and acknowledge that computations were in part supported by DFG-456666331.

## Author Contributions Statement

Z.L., S.T., and T.C.K. conceived and designed the experiments. Z.L. performed the experiments. Z.L., S.T. analyzed the data. Z.L., S.T., T.C.K contributed materials/analysis tools. Z.L., S.T., R.M. and T.C.K. wrote the paper. T.C.K. oversaw all parts of the research.

## Competing Interests Statement

The authors declare no competing interests.

## Supplementary Materials

| Model | α | β | λ |
|---|---|---|---|
| DVD-S | 1 | 1e-4 | 100 |
| DVD-B | 2 | 1e-4 | 150 |
| DVD-P | 4 | 4e-4 | 50 |

**Table S1 | Hyperparameters for the main DVD models.** *Each DVD variant is specified by a distinct combination of the hyperparameters $\alpha$, $\beta$, and $\lambda$. Respectively, these parameters govern: (i) the temporal mapping of development, (ii) the initial amplitude threshold in the frequency domain corresponding to contrast sensitivity at birth, and (iii) the mapping between threshold decay and the associated increase in contrast sensitivity over time. The reported values were selected through a hyperparameter sweep on mini-ecoset and subsequently applied to ecoset, ImageNet, and different architectures.*

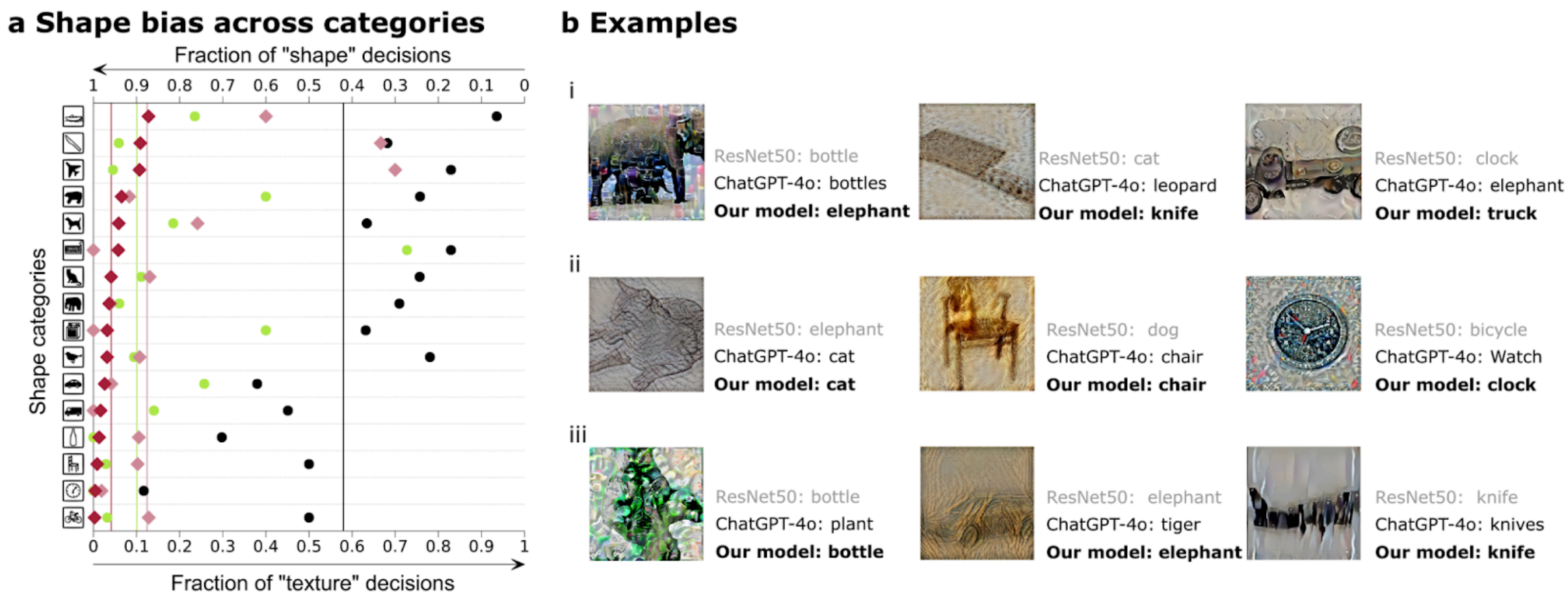

**Figure S1 | Category-specific shape bias and prediction examples. a.** *Category-wise analysis of shape bias in DVD-B (green) compared to humans (ages 4-6 and adult; pink and red) and the ResNet-50 baseline (black). Mean across seeds shown (n = 16; unit of analysis = category).* **b.** *Example predictions from different models on cue-conflict stimuli. Row 1: Only DVD-B makes shape-biased predictions, while both the ResNet-50 baseline model and ChatGPT-4o make texture-biased predictions. Row 2: Both DVD-B and ChatGPT-4o generate shape-biased predictions, while only the baseline model is texture-biased. Row 3: Challenging stimuli where all models make texture-based predictions.*

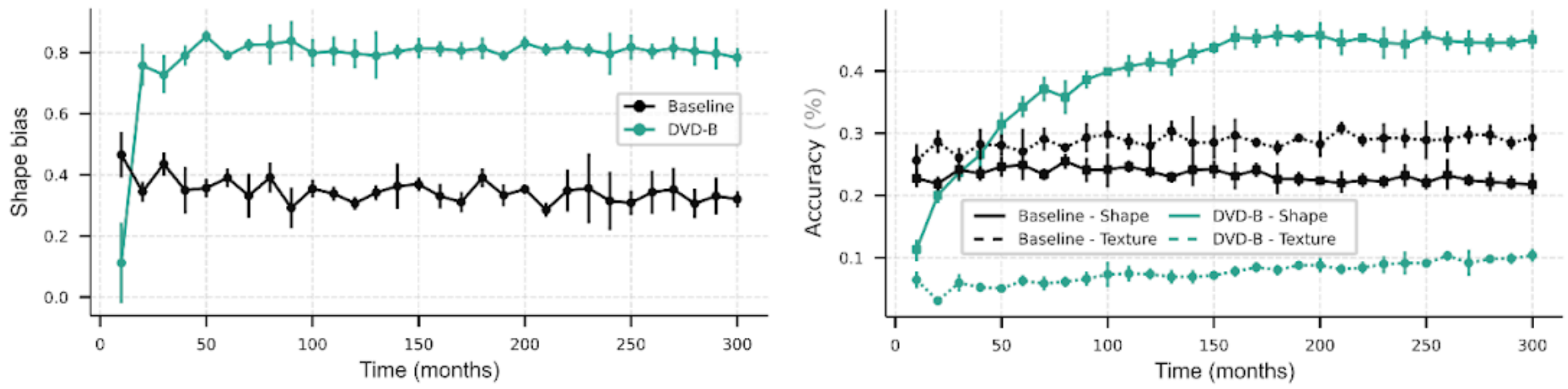

**Figure S2 | Evolving shape bias over development.** *Left: Evolving shape bias of DVD-B model from newborn to 25 years old (300 months, 2 months per epoch for DVD-B, shown at 10-months/5-epochs resolution), strong shape bias emerges very early by 20 months and gets maintained, in line with psychophysical evidence that babies are already shape biased at 2 years of age. Lines summarise shape bias across categories (n = 16; unit of analysis = category; centre = median across categories). Right: Percentage of shape-texture choices (correct predictions on shape/texture) across model training. Mean across independently trained model instances shown (n = 3 seeds); error bars indicate ±1 standard deviation across seeds*

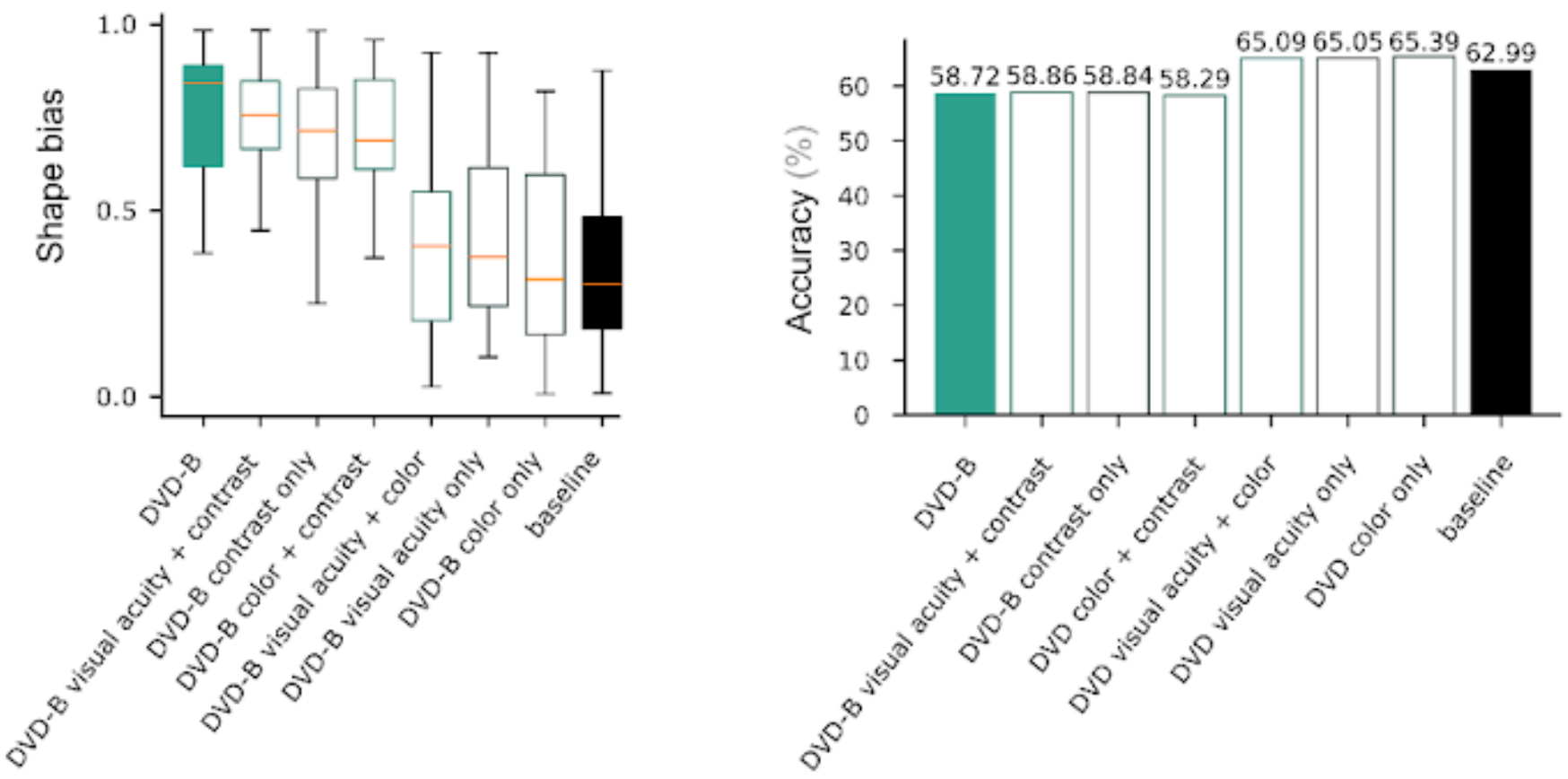

**Figure S3 | Controlled rearing to study the role of different visual aspects in development** (example: DVD-B)**.** *Left: Shape bias for all controlled-rearing settings, emphasising contrast development. Boxplots summarise category-wise shape bias across the 16 cue-conflict object categories (n = 16; unit of analysis = category); each condition corresponds to one independently trained model instance (one seed). Boxplot conventions are as in Fig. 2c. Right: Recognition performance of the controlled-rearing models. Bar plots show top-1 ecoset validation accuracy; each bar corresponds to one independently trained model instance (n = 1 seed per condition).*

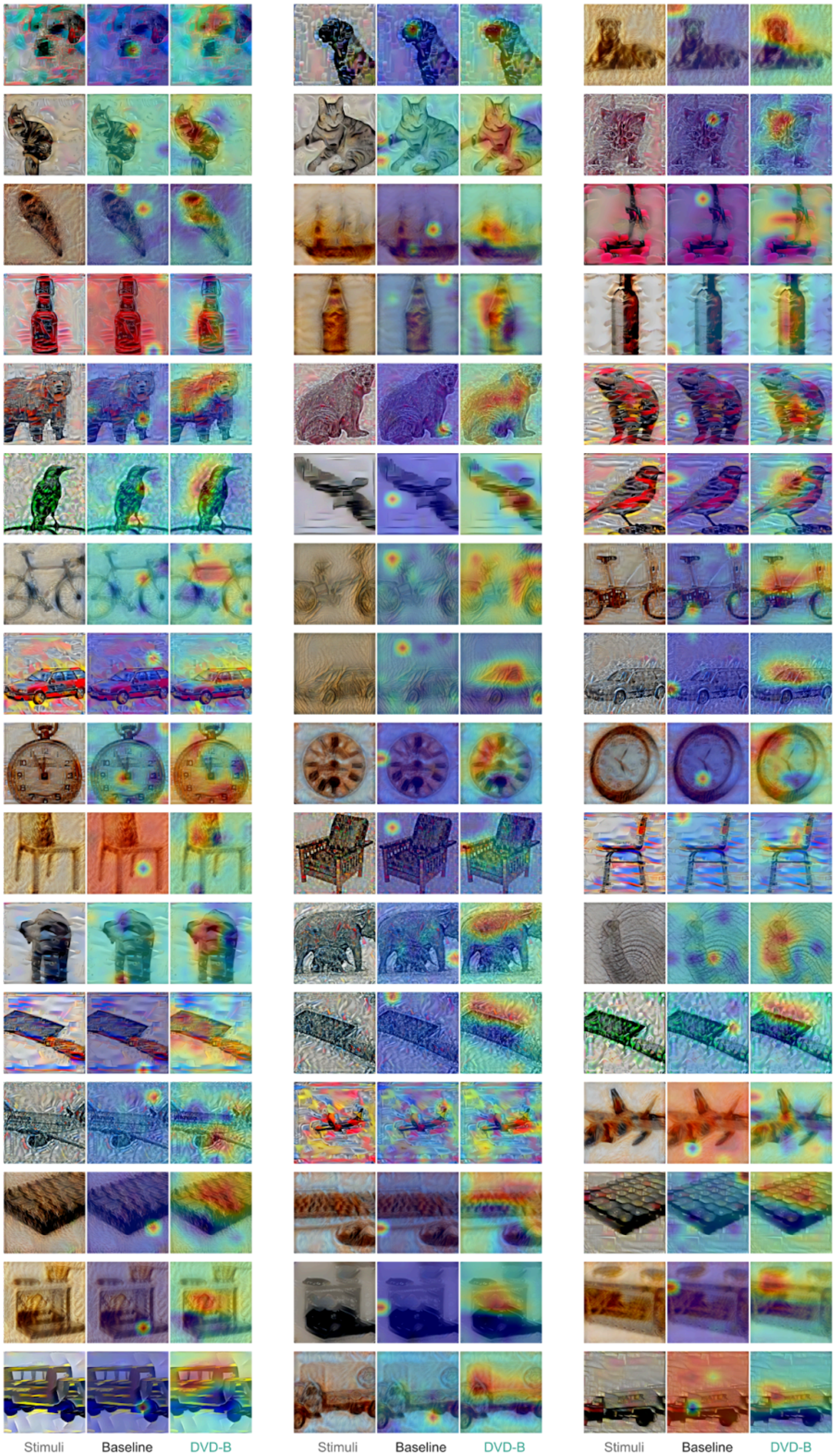

**Figure S4 | Feature importance visualisation using layer-wise relevance propagation**. *Additional examples illustrating the generality of the effects across stimuli and categories (n = 16 categories).*

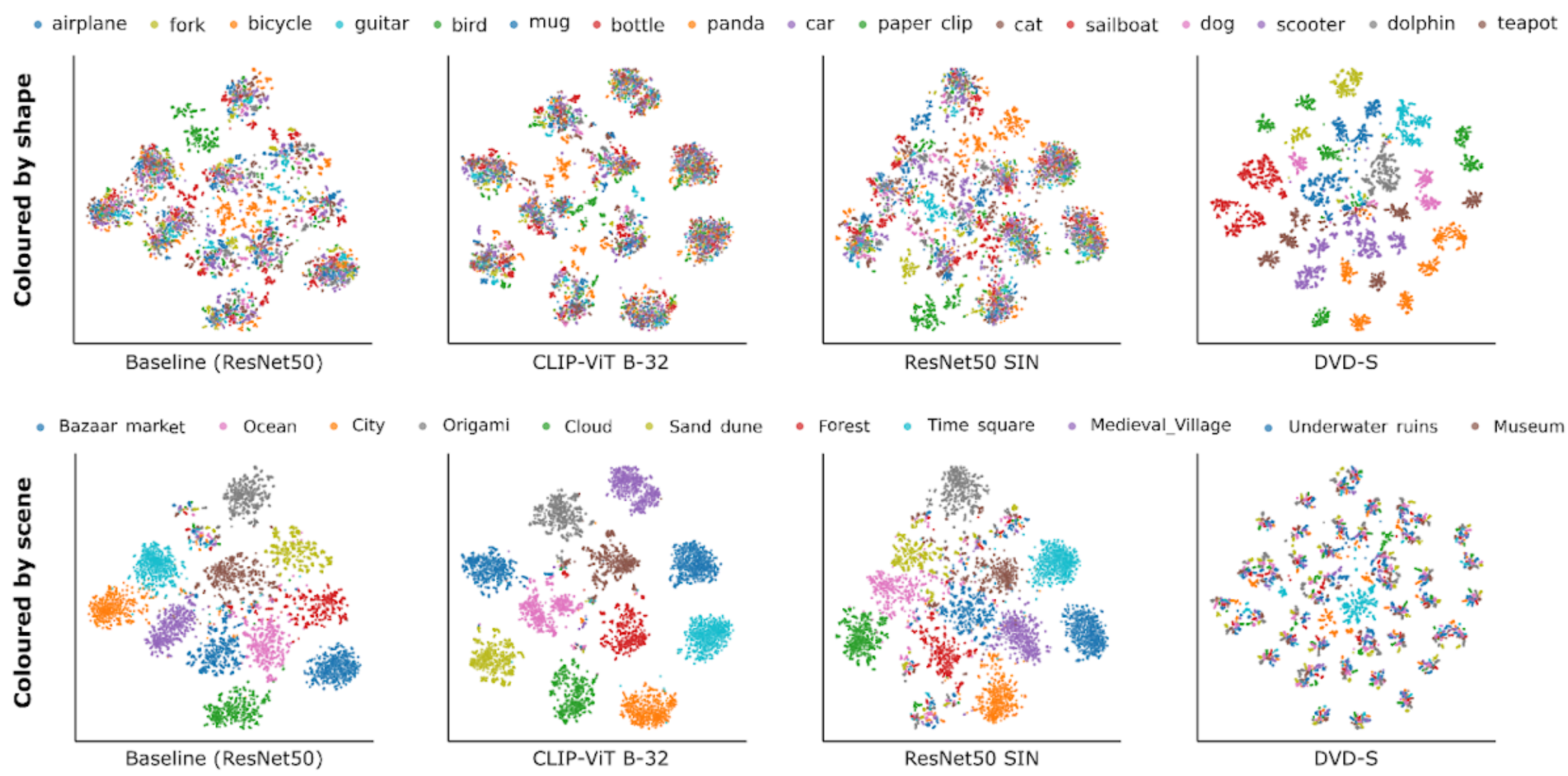

**Figure S5 | t-SNE visualisation distinguishing abstract shapes from scenes**. *t-SNE embeddings of IllusionBench images illustrate distinct clustering behaviour of abstract shapes under scene contexts for ResNet-50 baseline, CLIP-ViT B-32, ResNet-50 with style-transfer training, and DVD-trained models (DVD-S model shown as example). Each point corresponds to one image; colours denote the abstract shape category (top row) or the scene category (bottom row). Only DVD-trained models cluster images according to their abstract shape categories, whereas other models predominantly cluster images based on their scene contexts*

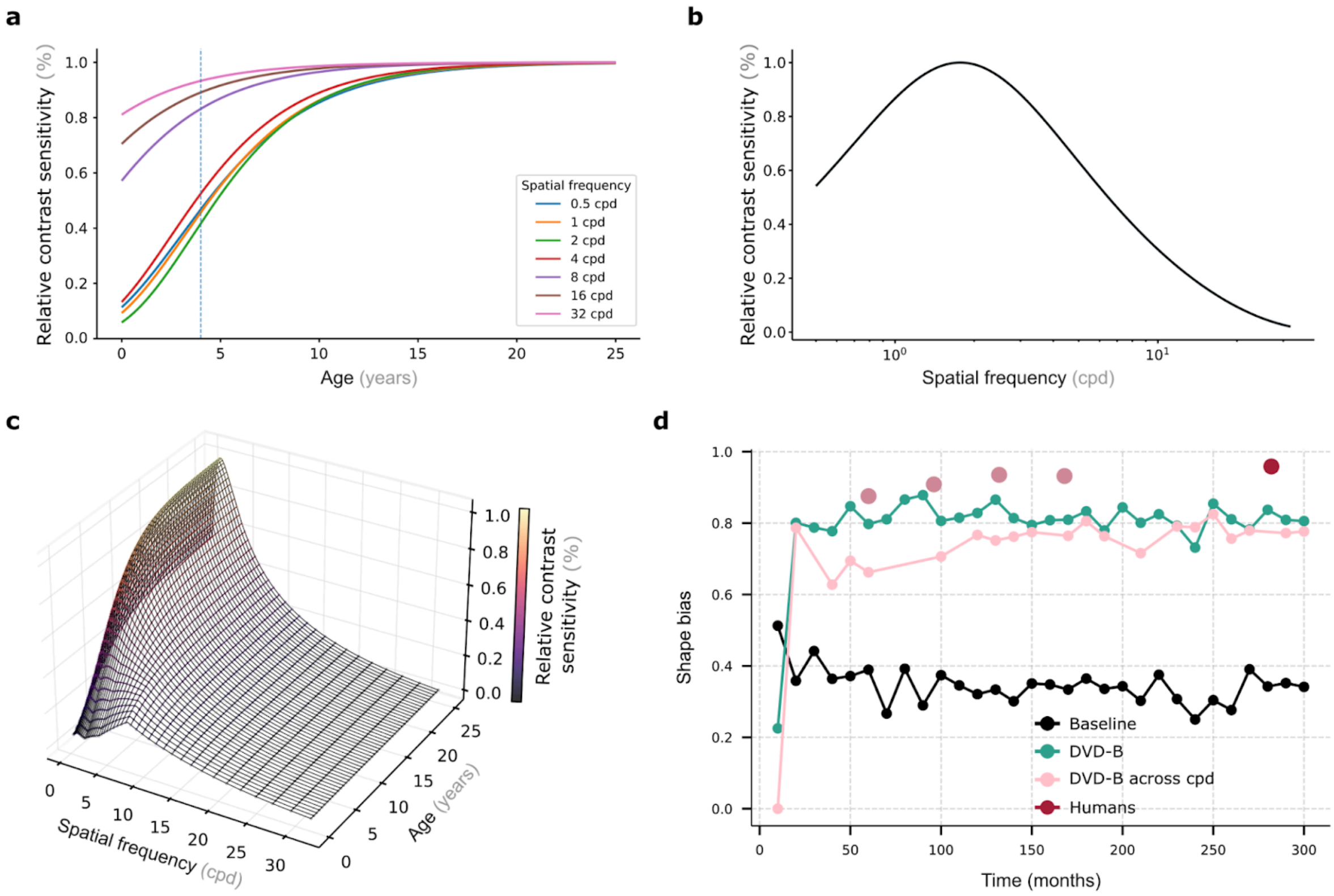

**Figure S6 | Modelling developmental contrast sensitivity across ages and spatial frequencies. a.** *Contrast sensitivity from newborns to adults varies across spatial frequencies (Dekker et al., 2020).* **b.** *Adult contrast sensitivity function (Barten et al., 1999)* **c.** *Modelling developmental contrast sensitivity (relative to adults) changes across both age and spatial frequency in a 3D mesh.* **d.** *Evolving shape bias of models from newborn to 25 years old (300 months, 2 months per epoch for DVD-B and DVD-B across cpd, shown at 10-months/5-epochs resolution), strong shape bias emerges very early both for DVD-B and DVD-B across spatial frequencies (cpd), and is largely maintained throughout training (n = 1 seed per condition). Lines summarise shape bias across categories (n = 16; unit of analysis = category; centre = median across categories).*

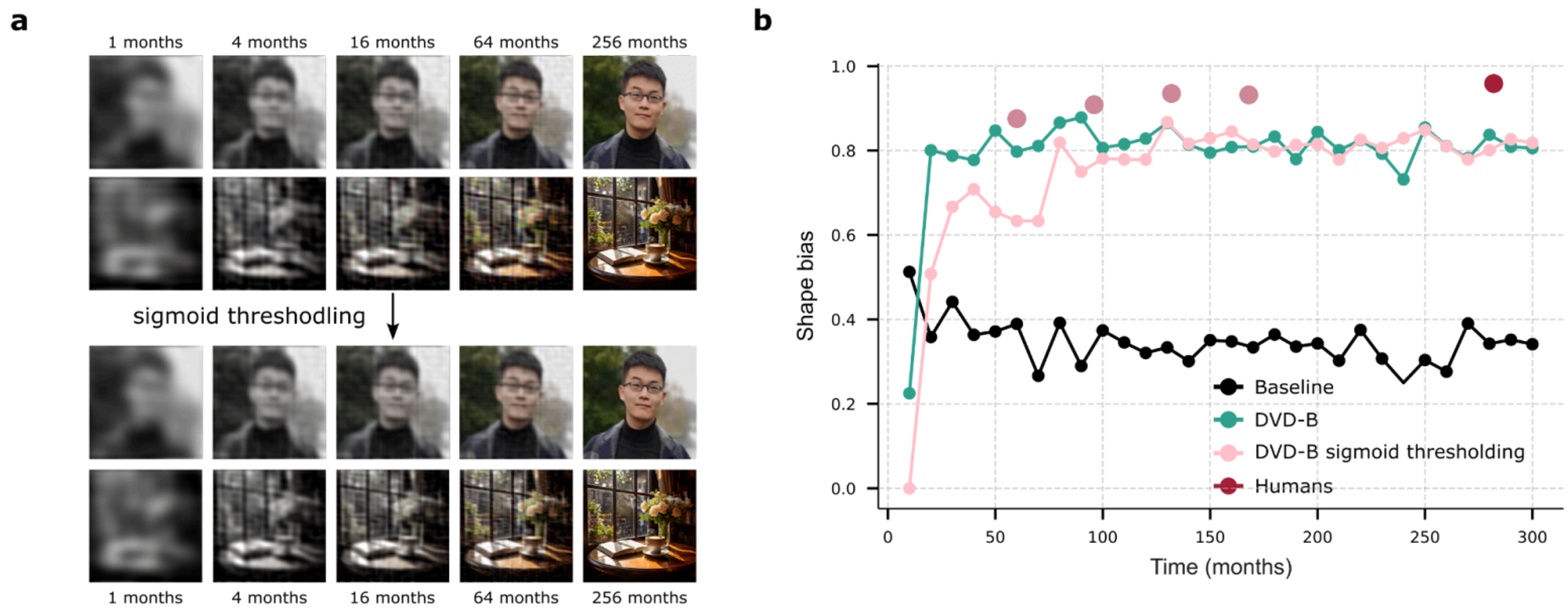

**Figure S7 | Sigmoid thresholding instead of hard thresholding in contrast sensitivity. a.** *Example images of DVD with sigmoid thresholding at selected developmental stages,* providing a smoother transition between visible and invisible contrasts compared to hard thresholding. **b.** *Evolving shape bias of models from newborn to 25 years old (300 months, 2 months per epoch for DVD-B and DVD-B sigmoid thresholding, shown at 10-months/5-epochs resolution), strong and similar shape bias emerges both for DVD-B and DVD-B sigmoid thresholding, and is largely maintained throughout training. Each curve corresponds to one independently trained model instance (n = 1 seed per condition). Median shape bias across categories shown (n = 16; unit of analysis = category).*

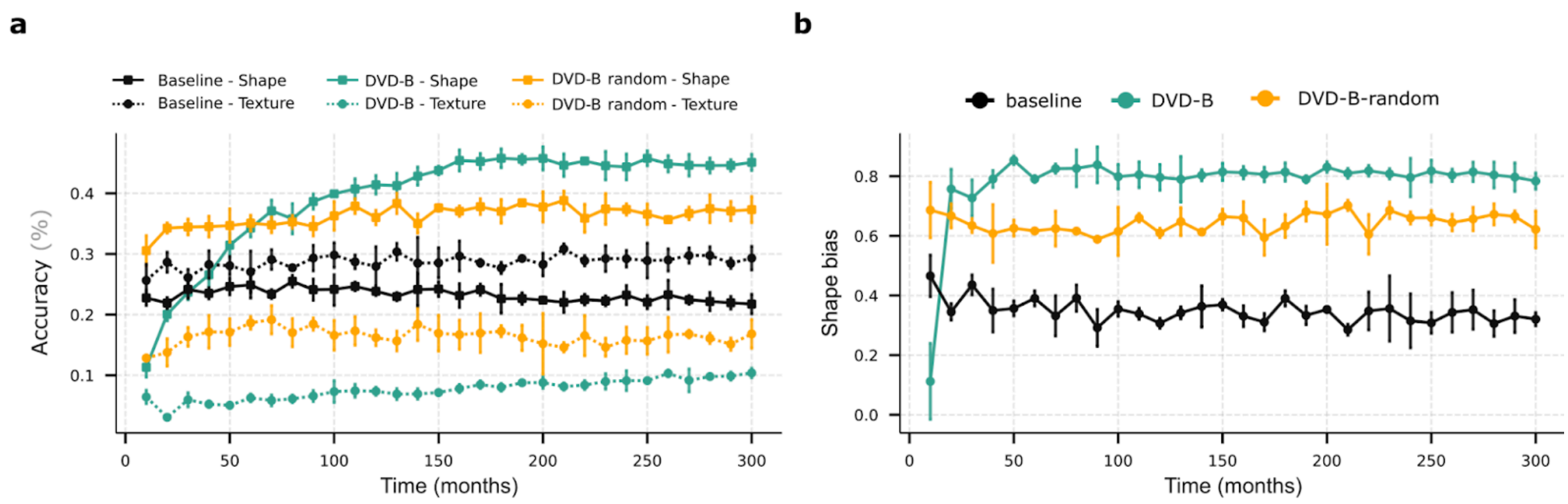

**Figure S8 | Randomly shuffled versus chronological developmental visual diet a.** *Percentage of shape-texture choices (correct predictions on shape/texture) across model training from newborn to 25 years old (300 months, 2 months per epoch for DVD-B, shown at 10-months/5-epochs resolution). Compared with randomly shuffled DVD-B, chronological DVD-B consistently made more shape choices and fewer texture choices during the middle and later phases of training.* **b**. *The evolving shape bias of chronological DVD-B remains consistently higher than that of randomly shuffled DVD-B after 20 months. Mean across seeds shown (n = 3); error bars indicate the ±1 standard deviation across seeds.*

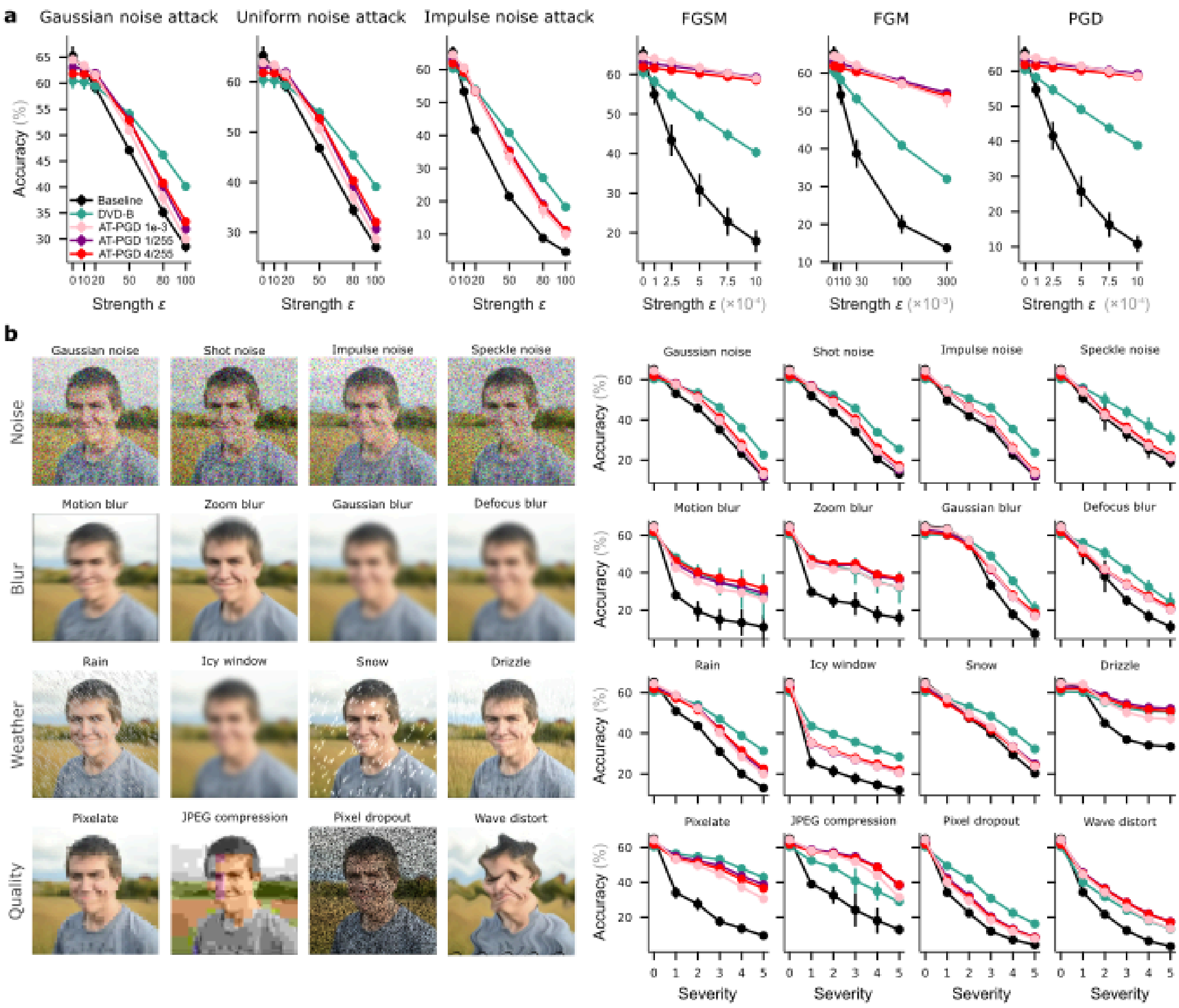

**Figure S9 | Comparison with an adversarially trained model. a.** *Robustness of DVD-B versus adversarially trained (PGD, ε=1e-3, 1/255 and 4/255) and baseline models under adversarial perturbations. White-box attacks include FGSM, FGM, and PGD; black-box attacks include Gaussian/uniform additive noise ($L_2$-norm) and impulse noise. While adversarial training confers resilience to white-box attacks, it generalizes poorly to black-box perturbations, performing worse than DVD trained model. This suggests that DVD training offers a more generalizable route to robustness, whereas adversarial training remains largely attack-specific.* **b**. *Performance of DVD-B, adversarially trained (PGD, ε=1e-3, 1/255 and 4/255), and baseline models under diverse naturalistic degradations. Adversarial training confers advantages for JPEG compression, yet DVD training yields superior or similar robustness across most other conditions. In a-b, mean across seeds shown (n = 3 seeds for baseline, DVD-B and PGD ε = 1e-3; n = 1 seed for PGD ε = 1/255 and 4/255); error bars indicate ±1 standard deviation across seeds .*

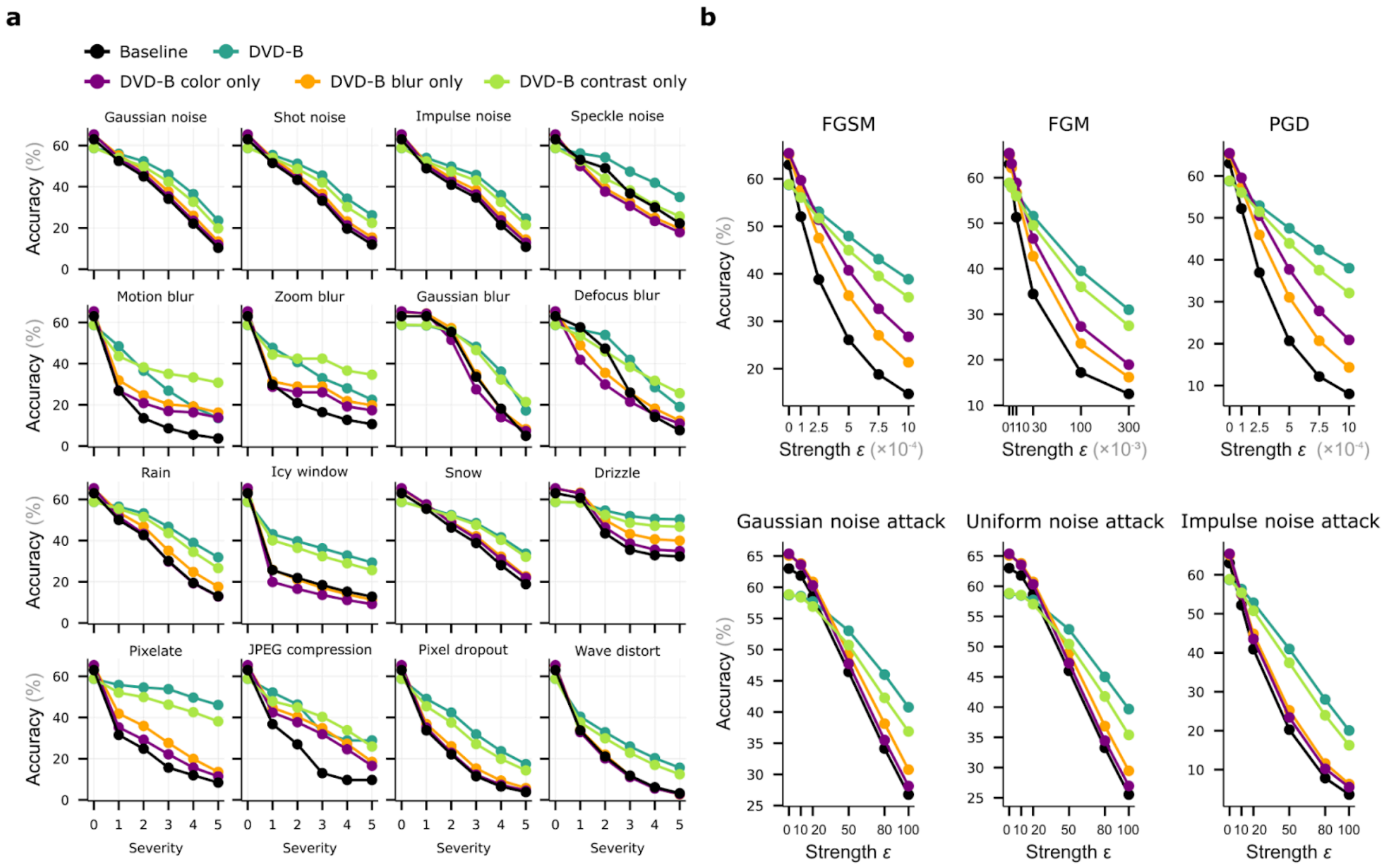

**Figure S10 | Analyses of the separate role of visual acuity, contrast, and colour for robustness. a.** *Robustness to naturalistic corruptions (noise, blur, weather, compression) in models trained solely with visual acuity, contrast, or colour (each condition: n = 1 independently trained model instance; one seed). Contrast sensitivity emerges as the dominant contributor across conditions. Acuity and colour afford selective gains under blur (motion, defocus) and JPEG compression, but are largely ineffective against noise (e.g., speckle) and weather-based corruptions. Integrating all three factors yields the highest overall resilience.* **b.** *Robustness of the models trained solely with visual acuity, contrast, or colour, to adversarial perturbations (each condition: n = 1 independently trained model instance; one seed). White-box attacks: FGSM, FGM, PGD. Black-box attacks: additive Gaussian/uniform noise ($L_2$-norm) and impulse noise. Contrast again provides the strongest defence relative to acuity or colour alone, though their combination achieves maximal robustness.*

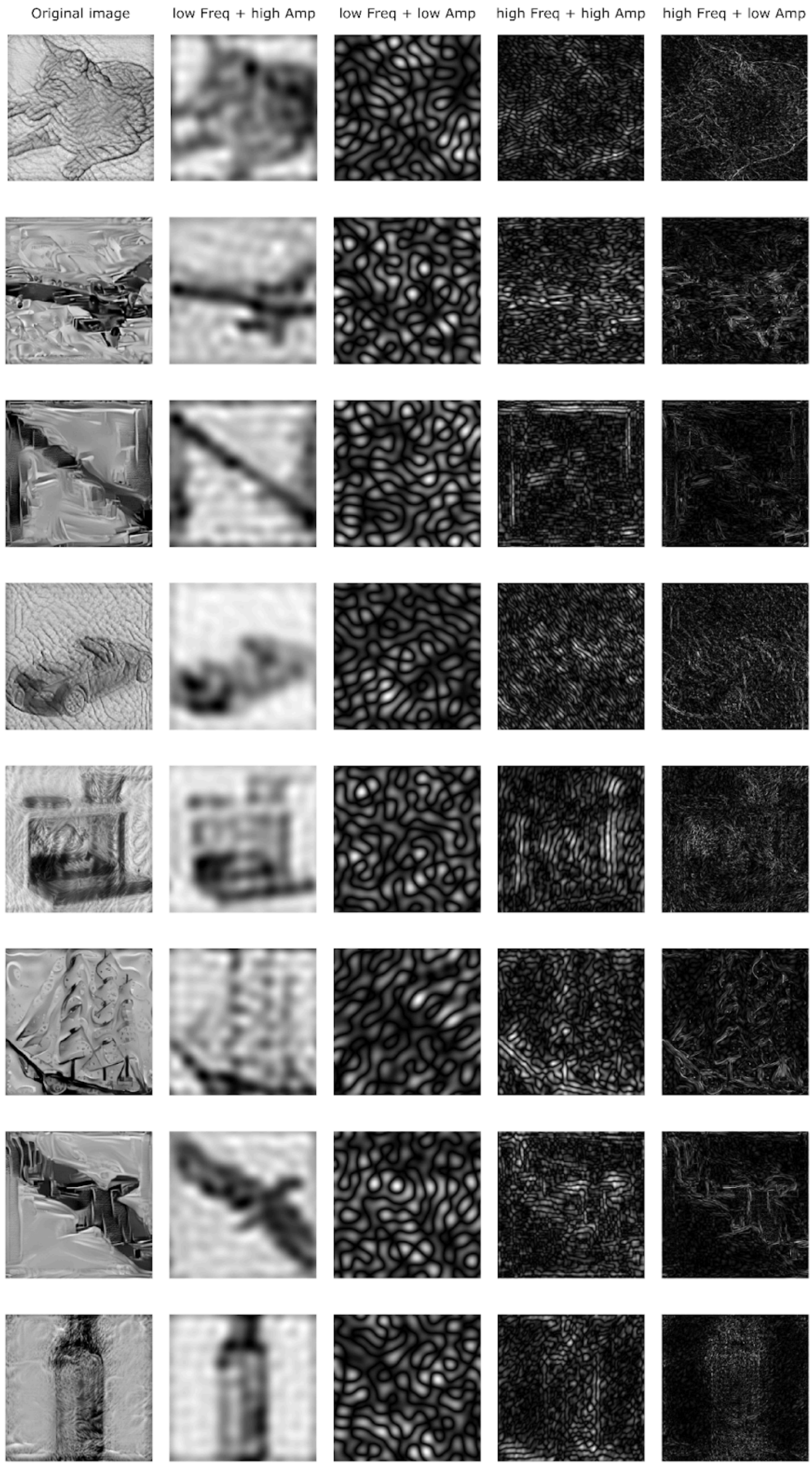

**Figure S11 | Contrast-dependent integrity of object structure.** *High-contrast low-frequency signals preserve global object structure (second column), whereas low-contrast counterparts degrade into*

*residual textures with minimal shape coherence (third column). These examples highlight contrast sensitivity as a critical source of spatially integrated shape information.*

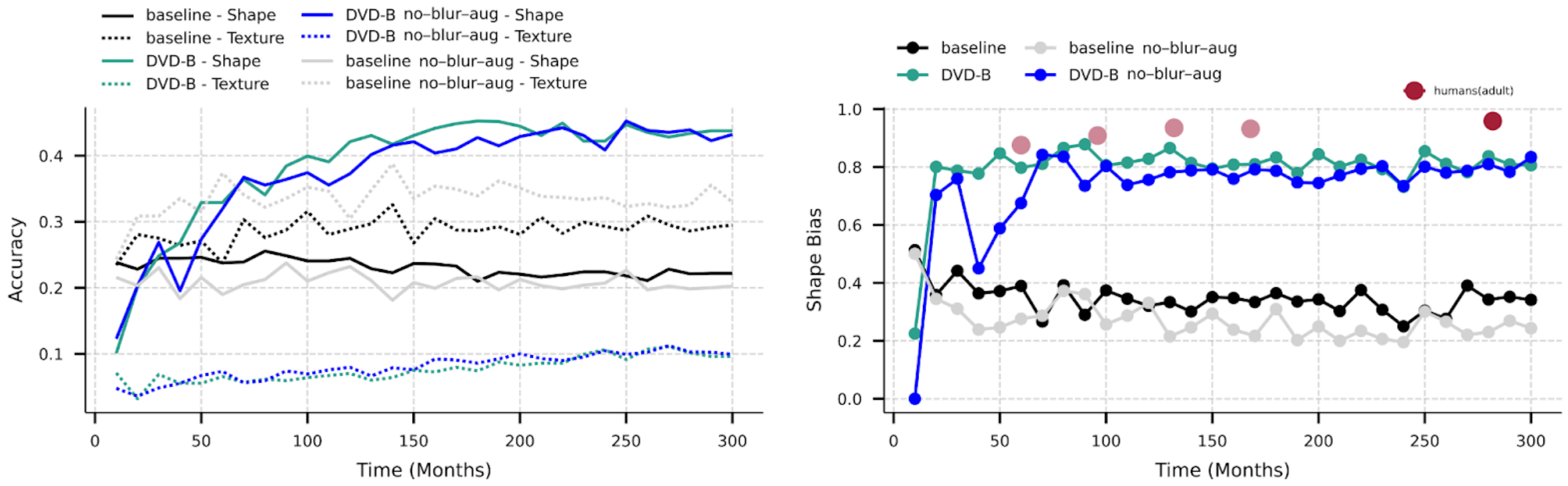

**Figure S12 | Control without Gaussian blur augmentation.** *Left: Percentage of shape-texture choices (correct predictions on shape/texture) across model training (n = 1 independently trained model instance; one seed). Right: Evolving shape bias of DVD-B model from newborn to 25 years old (300 months, 2 months per epoch for DVD-B, shown at 10-months/5-epochs resolution), strong shape bias emerges very early by 20 months and gets maintained, in line with psychophysical evidence that babies are already shape biased at 2 years of age. Shape bias outcomes remained consistent for a model trained without Gaussian blur data augmentation (n = 1 independently trained model instance; one seed). Median across categories shown (n = 16; unit of analysis = category).*

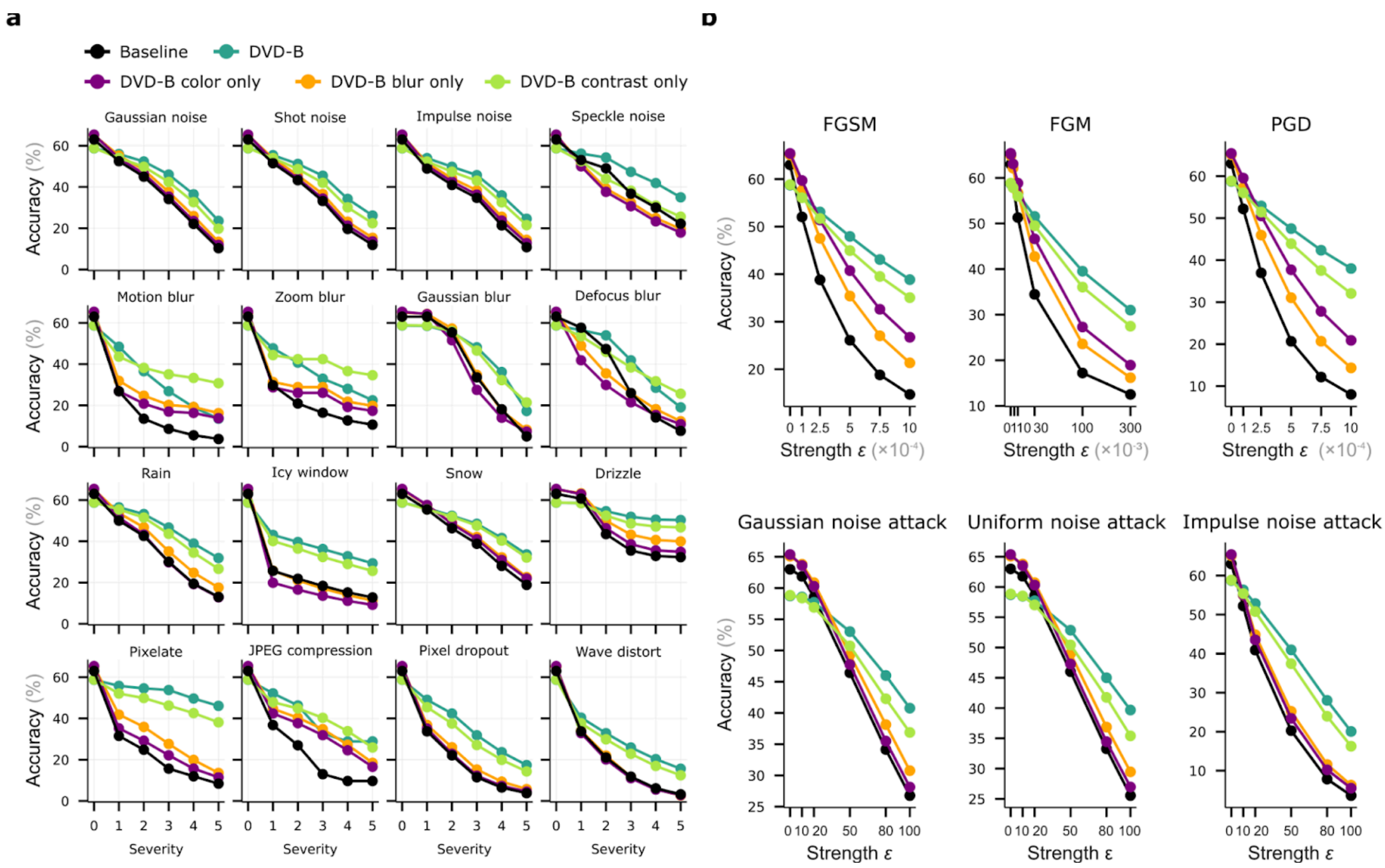

**Figure S13 | Robustness analyses across different DVD variants.** ***a.*** *Robustness to different naturalistic degradations, including noise, blur, weather, and distortion. DVD-trained models*

*outperform the baseline across all conditions, with performance systematically increasing from DVD-P to DVD-B and reaching the highest levels with DVD-S in most cases. Each curve corresponds to one independently trained model instance (n = 1 seed per model).* ***b.*** *Robustness to adversarial perturbations. DVD variants again show improved resilience relative to the baseline across both white-box (FGSM, FGM, PGD) and black-box (Gaussian, uniform, impulse noise) attacks, increasing from DVD-P to DVD-B and DVD-S (n = 1 independently trained model instance per model).*

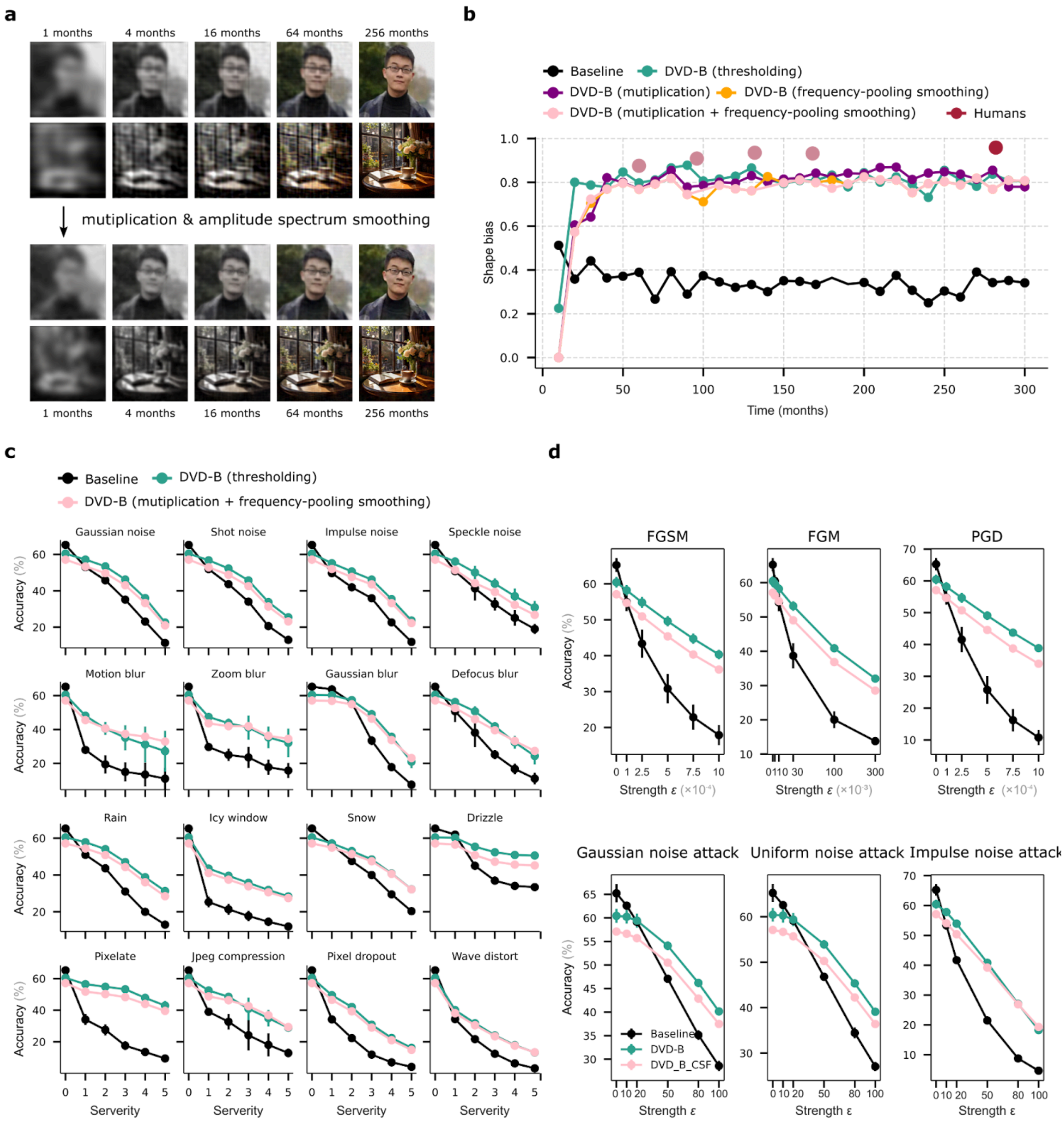

**Figure S14 | Contrast sensitivity via CSF multiplication and frequency pooling. a.** *Example DVD images after applying contrast sensitivity function (CSF) multiplication with frequency pooling at selected developmental stages, illustrating a smoother transition between visible and invisible contrasts than with hard CSF thresholding.* **b.** *Evolving shape bias of models from newborn to 25 years old (300 months, 2 months per epoch; plotted at 10-month/5-epoch resolution) for DVD-B with hard thresholding, CSF multiplication, frequency pooling, and their combination. Strong and similar shape bias emerges across all DVD-B and DVD-B variants and is largely maintained throughout training (n = 1 independently trained model instance; one seed). Median across categories shown (n = 16; unit of analysis = category).* **c.** *Robustness to naturalistic corruptions (noise, blur, weather, compression) in model trained with CSF multiplication and frequency pooling smoothing, showing robustness patterns consistent with the thresholding-based DVD approach relative to the baseline.* **d.** *Robustness of the model trained with CSF multiplication and frequency pooling smoothing to adversarial perturbations, with results similar to thresholding-based DVD-B relative to the baseline.*

*White-box attacks: FGSM, FGM, PGD. Black-box attacks: additive Gaussian/uniform noise ($L_2$-norm) and impulse noise. In c-d, mean across independently trained seeds shown (n = 3 seeds for DVD-B and baseline; n = 1 seed for model trained with CSF-multiplication and frequency-pooling); error bars indicate ±1 standard deviation across seeds .*

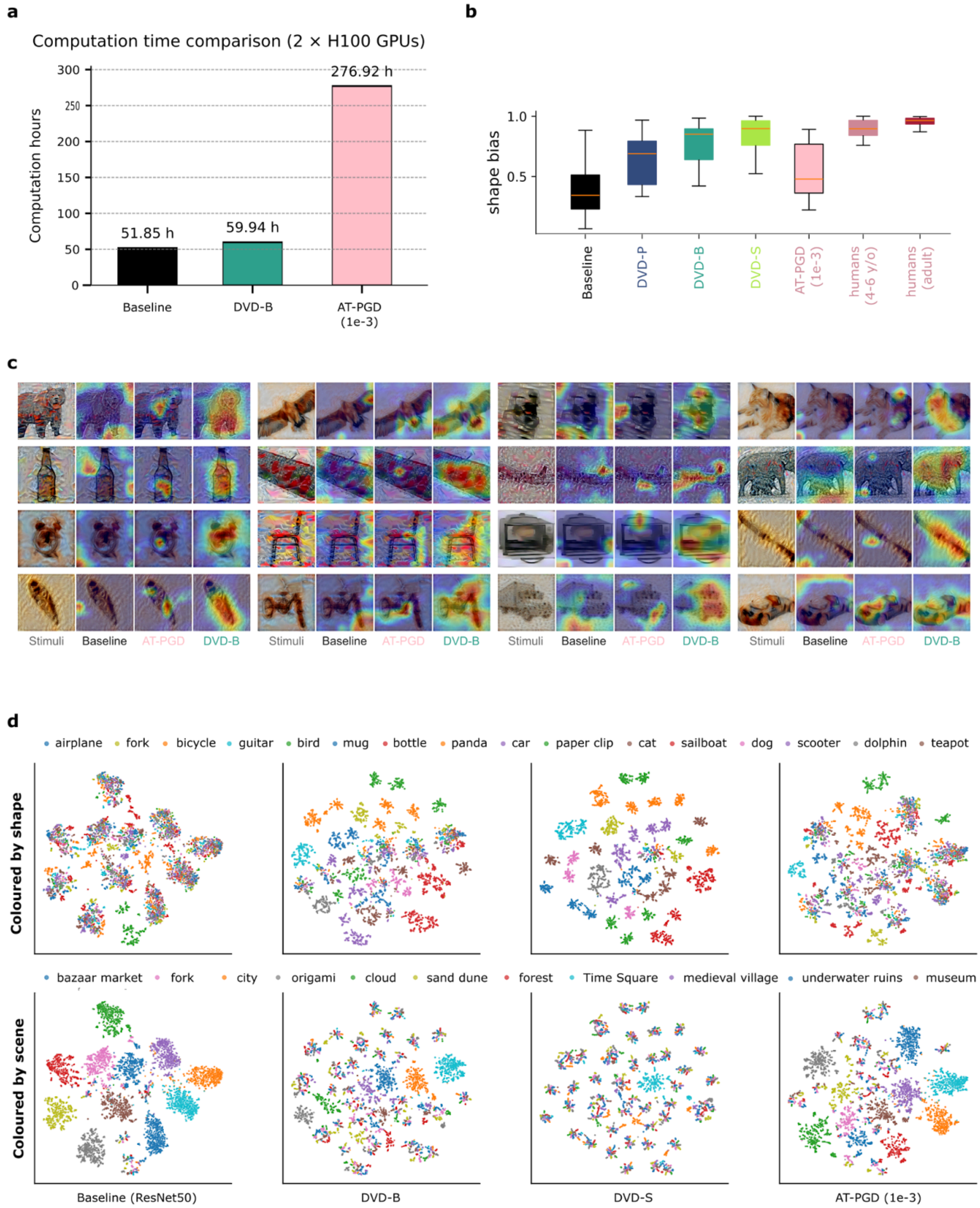

**Figure S15 | Comparisons of DVD-trained and adversarially trained models. a.** *Training-time comparison for the baseline, DVD-B, and adversarially trained (AT) models, showing that AT is substantially more time-consuming than both the baseline and DVD-trained models.* **b.** *Shape-bias comparison across the baseline, DVD variants, the AT model, and humans; the AT model exhibits lower shape bias than the DVD variants and remains well below the human range. Boxplots summarise category-wise shape bias across the 16 cue-conflict object categories (n = 16; unit of analysis = category) for each model (one seed; independently trained model instance). Human sample sizes and boxplot conventions are as in Fig. 2c.* **c.** *Grad-CAM feature attribution, indicating that DVD-B focuses more strongly on object-relevant regions than both the baseline and AT models.* **d.** *t-SNE visualization distinguishing abstract shapes from scenes. Embeddings of IllusionBench images illustrate different clustering patterns for the ResNet-50 baseline, DVD-B, DVD-S, and AT models. DVD-trained models cluster images according to abstract shape categories, whereas the baseline and AT models cluster primarily according to scene context. Each point corresponds to one image; colours denote the abstract shape category (top row) or the scene category (bottom row).*

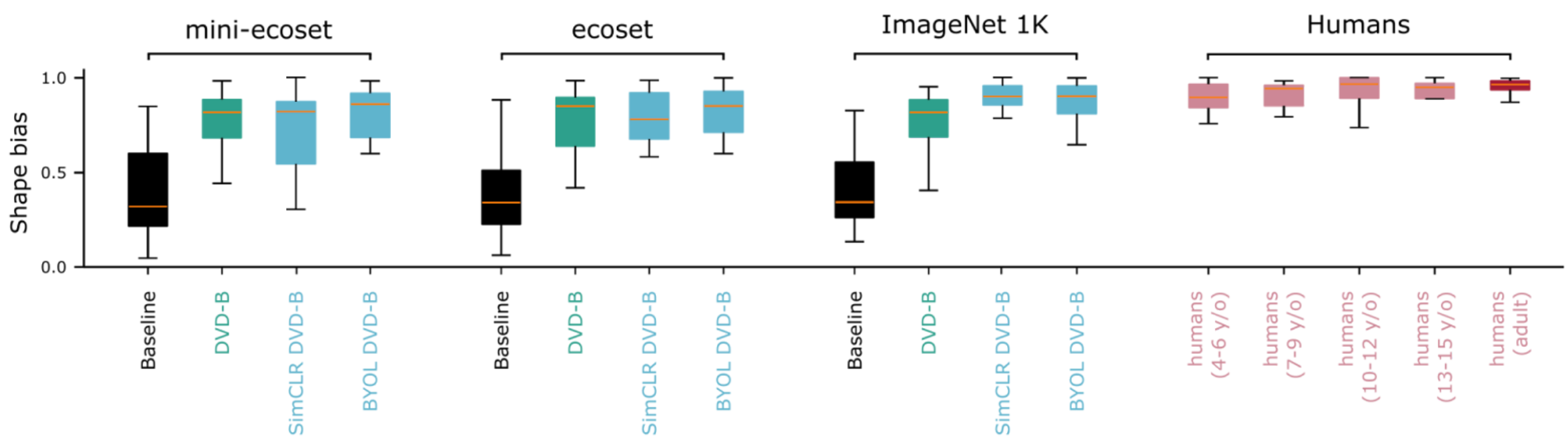

**Figure S16 | Shape-bias generalisation from supervised to self-supervised learning across datasets.** *Generalisation of shape bias results across training datasets (0.28 million mini-ecoset, 1.50 million ecoset, and 1.28 million ImageNet-1K) for self-supervised DVD-B models trained with SimCLR and BYOL. Boxplots summarise category-wise shape bias across the 16 cue-conflict object categories (n = 16; unit of analysis = category) for each model (one seed; independently trained model instance). Human sample sizes and boxplot conventions are as in Fig. 2c.*